\documentclass[lettersize,journal]{IEEEtran}
\usepackage{amsmath,amsfonts}
\usepackage{algorithmic}
\usepackage{algorithm}
\usepackage{array}
\usepackage[caption=false,font=footnotesize,labelfont=rm,textfont=rm]{subfig}
\usepackage{textcomp}
\usepackage{stfloats}
\usepackage{url}
\usepackage{verbatim}
\usepackage{graphicx}
\usepackage{cite}
\usepackage{widetext}
\usepackage{amssymb}
\usepackage{booktabs}
\usepackage{color, colortbl}
\usepackage[font=small]{caption}

\begin{document}

\title{A Safe and Efficient Self-evolving Algorithm for Decision-making and Control of Autonomous Driving Systems}

\author{Shuo Yang, Liwen Wang, Yanjun Huang$^*$, Hong Chen,~\IEEEmembership{Fellow,~IEEE}


\thanks{

Shuo Yang and Liwen Wang are with School of Automotive studies, Tongji University, Shanghai 201804, China (e-mail:shuo\_yang@tongji.edu.cn, 2131525@tongji.edu.cn)

Yanjun Huang is with School of Automotive studies, Tongji University, Shanghai 201804, China, and is also with Frontiers Science Center for Intelligent Autonomous Systems,
Shanghai 200120, China (corresponding author: e-mail: yanjun\_huang@tongji.edu.cn)

Hong Chen is with the College of Electronics and Information Engineering, Tongji University, Shanghai 201804, China (e-mail: chenhong2019@tongji.edu.cn)

}
}



\maketitle

\begin{abstract}

Autonomous vehicles with a self-evolving ability are expected to cope with unknown scenarios in the real-world environment. Take advantage of trial and error mechanism, reinforcement learning is able to self evolve by learning the optimal policy, and it is particularly well suitable for solving decision-making problems. However, reinforcement learning suffers from safety issues and low learning efficiency, especially in the continuous action space. Therefore, the motivation of this paper is to address the above problem by proposing a hybrid Mechanism-Experience-Learning augmented approach. Specifically, to realize the efficient self-evolution, the driving tendency by analogy with human driving experience is proposed to reduce the search space of the autonomous driving problem, while the constrained optimization problem based on a mechanistic model is designed to ensure safety during the self-evolving process. Experimental results show that the proposed method is capable of generating safe and reasonable actions in various complex scenarios, improving the performance of the autonomous driving system. Compared to conventional reinforcement learning, the safety and efficiency of the proposed algorithm are greatly improved. The training process is collision-free, and the training time is equivalent to less than 10 minutes in the real world.

\end{abstract}

\begin{IEEEkeywords}
Autonomous driving, Reinforcement learning, Self-evolving
\end{IEEEkeywords}

\section{Introduction}
\IEEEPARstart{w}{ith} the development of advanced technologies such as computer science and artificial intelligence, autonomous driving technology is attracting more and more attention \cite{tan2024integrating}. Autonomous vehicles are expected to move occupants from one location to another, while reduce costs, improve safety, and reduce traffic congestion \cite{jacobstein2019autonomous}. As the "brain" of the autonomous vehicles, the decision-making and control is required to fully consider the interaction with other traffic participants and generate appropriate actions to complete driving tasks. Its capability directly reflects the intelligence of the autonomous driving system\cite{chai2022deep}.

Rule-based methods are the most widely used at present for decision making and control. 
By analyzing the behavioral characteristics and decision logic of human drivers, engineers extract and design the corresponding rules. Common rule-based methods include finite state machine \cite{kammel2008team}\cite{ziegler2014making}, sampling method \cite{zhang2019trajectory}, etc. The advantage of these methods is that their output is known explicitly in the case of any given information with good interpretability. However, they rely on a large amount of expert knowledge and manually designed logic, which is difficult to cover all complex scenarios.

In recent years, several studies have applied optimization-based methods. These methods abstract the trajectory planning for autonomous driving as an optimization problem to find the optimal behavior under a certain optimization objective \cite{zhou2023event}\cite{stano2023model}\cite{Huang_TVT}. Nilsson et al. \cite{nilsson2013strategic} proposed an overtaking decision method, which modeled the mixed logic dynamic system and realized the optimal solution based on the mixed integer programming. Farkas et al. \cite{farkas2022mpc} introduced a model predictive control with a centralized controller to realize speed planning, so as to reduce the blockage in roundabouts. The optimization-based methods can obtain the optimal solution in a certain traffic scenario. However, usually it is difficult to construct optimization problems when dealing with both decision-making and control as a whole.

Deep Reinforcement Learning (DRL) is a recent approach that combines the learning ability of deep learning with the decision-making abilities of reinforcement learning \cite{valiente2023prediction}\cite{sewak2019deep}. Many studies have applied DRL to decision-making algorithms in autonomous driving, showing obvious performance improvements. Liu et al. \cite{liu2024augmenting} proposed  the Scene-Rep Transformer to enhance RL decision-making  capabilities through improved scene representation encoding and sequential predictive latent distillation. The proposed method is validated in a challenging dense traffic simulated city scenario. Yang et al. \cite{yang2023towards} proposed a robust decision-making framework for highway autonomous driving based on DRL. When encountering unseen scenarios, the rule-based algorithm will take over the learning based policy to ensure the lower limit performance. He et al. \cite{he2024trustworthy}proposes a new defense-aware robust RL approach designed to ensure the robustness of autonomous vehicles in the face of worst-case attacks. Simulation and experimental results show that the proposed method can still provide a credible driving policy even in the presence of the worst-case observed disturbance. It can be seen that compared to rule-based and optimization-based methods, such algorithms can realize self-training via interacting with the environment, and has the potential to deal with more complex and varying traffic environment.

However, although RL algorithm play an important role in decision-making and control system for autonomous driving, they need to be trained in the simulation environment for hundreds of thousands or even millions of steps \cite{Prathiba_RL}, which is too inefficient to be applied. In addition, its performance heavily depends on the exploration of the environment, which makes it difficult to guarantee safety\cite{SONY_RL_Racing}. Obviously, it is impossible to apply to safety-critical systems \cite{peng2022model}\cite{lin2024safety}.

Therefore, the motivation of this paper is to propose a hybrid Mechanism-Experience-Learning augmented approach for decision-making and control of autonomous vehicles to solve the safety issues and low learning efficiency of RL algorithms. Aiming at safety issue, the constrained nonlinear optimization problem is constructed as the mechanism model to ensure the safety of exploration. To solve the inefficiency problem, the driving tendency is defined by analogy with human driving experience, so as to reduce the search space of autonomous driving problem and improve the learning efficiency. The classical actor critic architecture\cite{konda1999actor} in RL community is used to achieve the algorithm evolution, where the neural network and the nonlinear optimization are combined as the approximator of the actor function. The approach combines the advantages of rule-based, optimization-based, and learning based methods, taking traffic information as input and directly outputting the vehicle control actions. The contributions of this study are summarized as follows:

1) A hybrid approach based on Mechanical-Experience-Learning is proposed to solve the problems of poor safety and low efficiency of RL algorithms.

2) A concept of driving tendency is defined by analogy with human driving experience, to reduce the search space of RL problem and improve the learning efficiency.

3) An interaction mechanism model is constructed to determine the initial condition that is closed to the optimal solution to speed up policy update of the global optimization problem.

This paper is organized as follows: Section II introduces the proposed framework. The details of agent implementation are introduced in Section III. The traffic interaction model is described in Section IV. The design of safety evolutionary algorithm is described in Section V. In Section VI, the method is validated and analyzed under a complex highway scenario, and Section VII concludes this paper.

\section{Proposed Framework}

\begin{figure*}[ht!]
\centering
\begin{tabular}{c}
\includegraphics[width=0.65\textwidth]{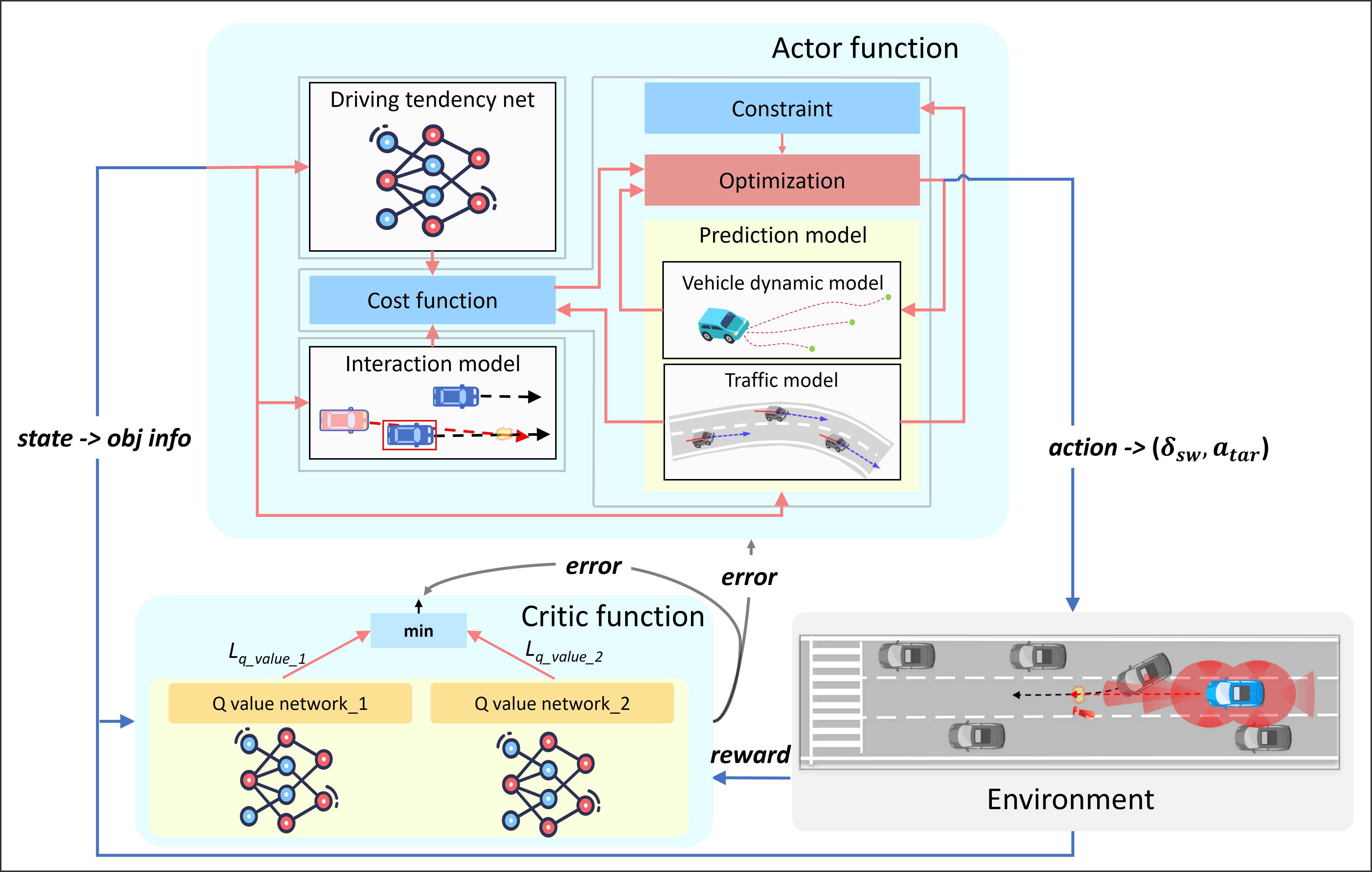}
\end{tabular}
\caption{Overall architecture of safe and efficient self-evolving algorithm for decision making and control.}
\label{overall architecture of integrated framework}
\end{figure*}

The proposed algorithm for decision-making and control of autonomous driving with self-evolutionary capabilities aims to generate efficient and safety control policy. It could not only provide reasonable actions in real time in response to changes in the complex traffic environment, but also allows for self-evolution during operation in the context of safe exploration. The algorithm is based on a typical actor-critic architecture that treats as a unified optimization problem to be solved. The main components of the proposed architecture, including the critic function approximator and the policy function approximator, are depicted in Fig. \ref{overall architecture of integrated framework}.

In this framework, the critic function approximator is used to estimate the value of the current policy and evaluate the effectiveness of the policy. The value function is chosen to consist of two Q-value networks \cite{haarnoja1861soft}, and it estimates the value by selecting the minimum value between the two Q-value networks $Q_1$ and $Q_2$, i.e., $min(Q_1, Q_2)$ . This approach is used to make the estimated value function closer to the true value in order to avoid overestimation. The policy approximator is employed to learn a safety policy with the objective of maximizing the return as much as possible. The policy approximator consists of three components, including an interaction model, a driver tendency network, and an optimization problem solver. A mechanism-based interaction model is applied to determine appropriate initial values for optimization and to achieve a hot start of the algorithm. In analogy to the driving process of a human driver, the concept of driving tendency is defined. It is used to quantifies the driving inclination during the driving process, which is a somewhat vague definition. The driving tendency network is designed to continuously improve its output, aiming to provide a more rational driving tendency that can effectively guide the optimization process of the algorithm. The resulting output, along with the output of the interaction model, is then given into the cost function. This cost function solves a nonlinear optimization problem, enabling it to generate safe and reliable lateral and longitudinal control actions for the vehicle.

\section{Agent implementation}

\subsection{Markov Decision Process (MDP)}

The autonomous driving problem can be seen as a Markov decision problem, which is defined as a tuple: $\left( {S,A,P,R} \right)\ $, where $S$ is state space, $A$ is action space, $P$ is transition model and $R$ is reward function. We define the input to the algorithm as $S$  and the output as $A$ .

Policy $\pi (s)$ is defined as a function of state to action. The optimization goal of the policy is to get an action that maximizes the expected reward. The expected rewards can be defined as:

\begin{equation}
\label{optimal policy}
\begin{aligned}
{v_\pi }(s) &= {_\pi }\left[ {{G_t}\mid {S_t} = s} \right]\\
& ={ }_\pi\left[\sum_{k=0}^{\infty} \gamma^k R_{t+k+1} \mid S_t=s\right]\ ,for\ all\ s \in {\cal S},
\end{aligned}
\end{equation}

where $G_t$ is the discounted cumulative reward, $S_t$ is the state at step $t$ and $\gamma $  is discounted factor.

The Q function is used to map the state-action pair to the expected return value and can be used to estimate the long-term cumulative return value that can be obtained by taking an action in a certain state. The Q function can be defined as:

\begin{equation}
\label{Q function}
\begin{aligned}
{q_\pi }(s,a) &= {_\pi }\left[ {{G_t}\mid {S_t} = s,{A_t} = a} \right] \\
&={_\pi }\left[ {\sum_{k=0}^{\infty} \gamma^k R_{t+k+1} \mid {S_t} = s,{A_t} = a} \right].\
\end{aligned}
\end{equation}

To obtain the maximum return, we can find the optimal policy by maximizing the value function. This can be achieved by using the Bellman optimality equation, which is defined as follows:

\begin{equation}
{v^\pi }(s) = \mathop {\max }\limits_{a \in {\cal A}} {q^\pi }(s,a).\
\end{equation}

\subsection{State design}

In this paper, the state space is designed to represent the observed environmental information.  The state input must encompass sufficient information of the agent to comprehend all crucial factors within the environment. It is essential to include important information in the state representation to enable the agent to make optimal decisions. To achieve this, the state space $s^f$ can be designed as:

\begin{equation}
\begin{aligned}
{s^f} = &[{x_e},y{}_e,{\varphi _e},{v_e},{x_1},{y_1},{\varphi _1},{v_1},{x_2},{y_2},{\varphi _2},\\
&{v_2},...,{x_n},{y_n},{\varphi _n},{v_n}],\
\end{aligned}
\end{equation}
where $n$ is the number of traffic vehicles in the observation area; $x_e$, $y_e$ are the lateral and longitudinal coordinates of the ego car in the global coordinate system; $\varphi _e$ is the heading angle of ego vehicle; $v_e$ is  the ego vehicle’s speed;  $x_i$, $y_i$ are the lateral and longitudinal coordinates of the traffic vehicles in the global coordinate system; $ \varphi _i$  are the heading angles of traffic vehicles; $v_i$ are  the traffic vehicles’ speed. All  $i=1,...,n$.

\section{Traffic Interaction Model Descriptions}

\begin{figure}[!t]
\centering
\includegraphics[width=3.5in]{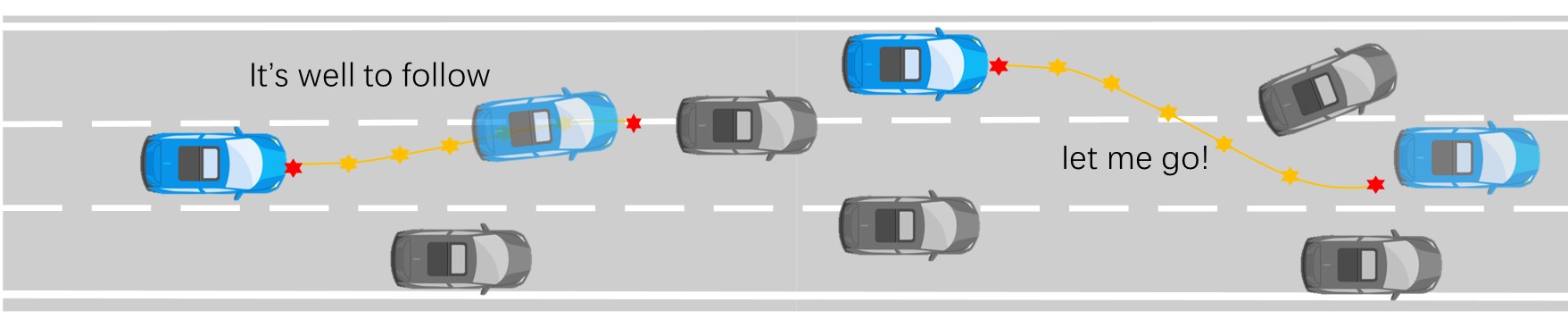}
\caption{Operating logic of driving tendency.}
\label{DT}
\end{figure}

For human drivers, there are two subconscious considerations during the driving process: what actions should I take, and what kind of interactions will occur with other vehicles in response to my actions. This involves a process of modeling, predicting, and deducing. More experienced drivers are better able to model their surrounding environment, anticipate the behavior of other vehicles, and generate more reasonable actions in response.

Analysis of driving psychology of the human driver shows that humans will create a target tendency at each moment of the driving process. Further, the human will estimate how the interaction between the surrounding traffic vehicles and the ego vehicle will be influenced by this driving tendency. Drawing an analogy to the cognitive process of human drivers, this section introduces the concept of "driving tendency". It quantifies the driving inclination during the driving process, which is a somewhat vague definition. In this section, a traffic interaction model under driving tendency is established for algorithm development, and a potential impact vehicles selection model is proposed to improve computational efficiency.

\subsection{Driving Tendency}

\begin{figure}[!t]
\centering
\subfloat[]{\includegraphics[width=2.5in]{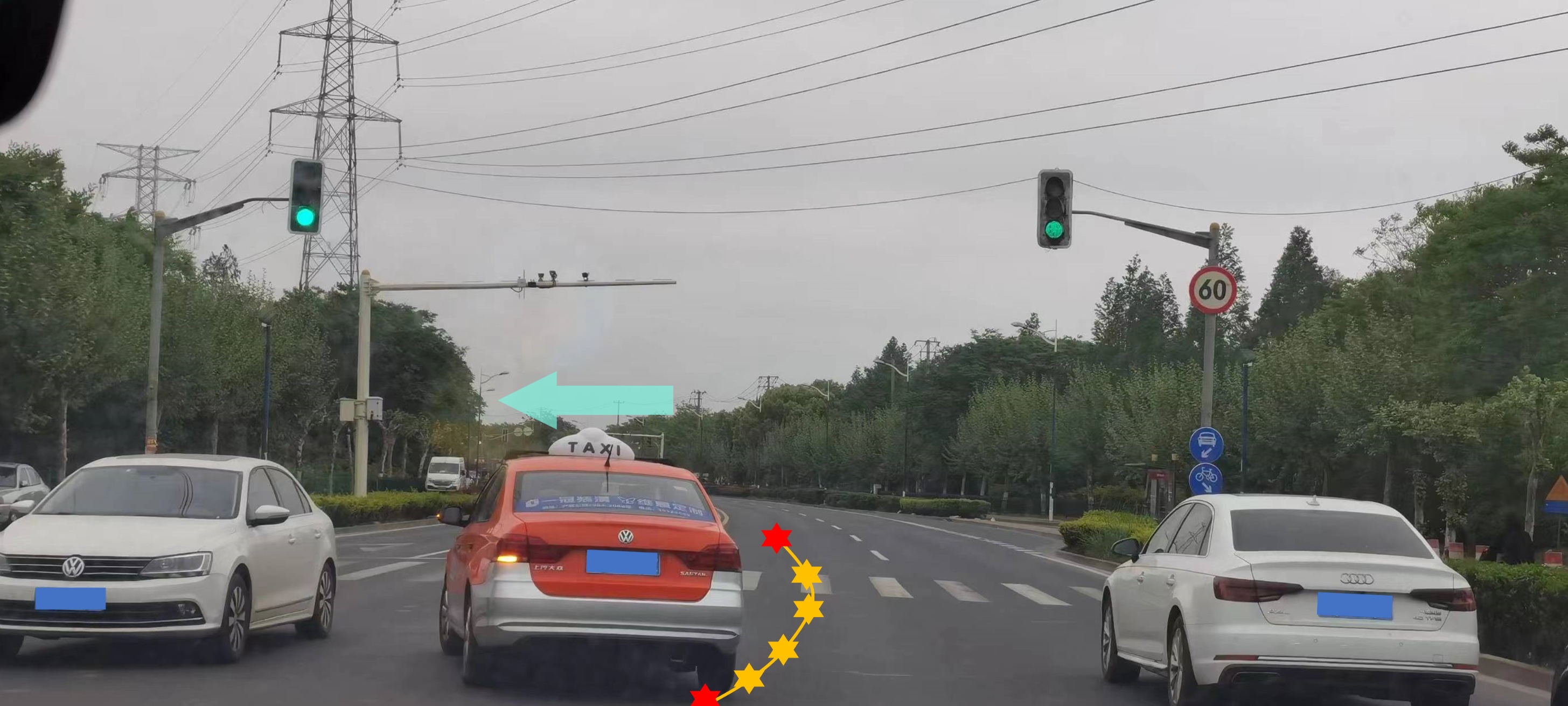}%
\label{DTexample_a}}
\hfil
\subfloat[]{\includegraphics[width=2.5in]{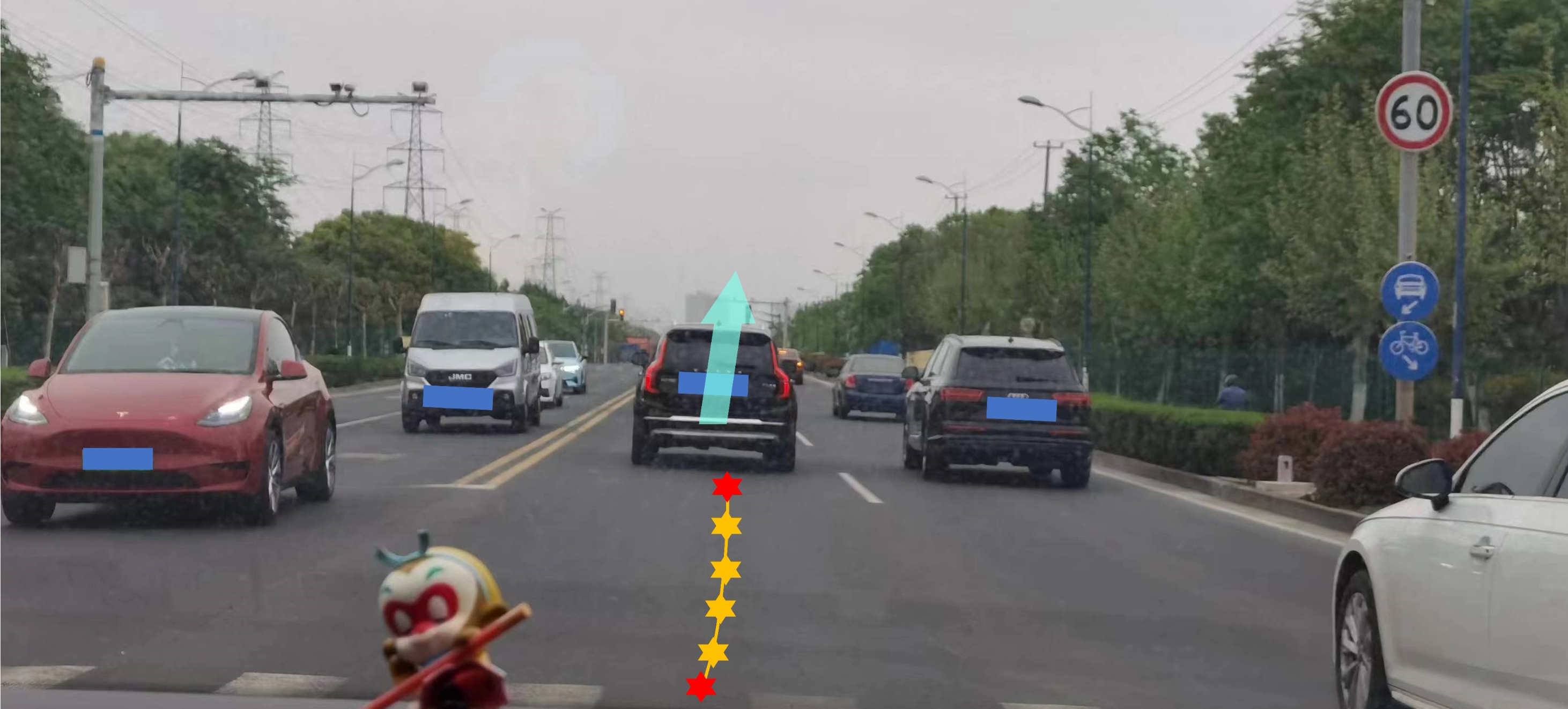}%
\label{DT_example_b}}
\caption{Schematic diagram of driving tendency in different scenarios}
\label{fig_4}
\end{figure}

First, analysis of the driving process considering human experience is presented. The driver usually gives a control command at each moment based on the actual state of surrounding vehicles. In this process, the driver has a vague idea, for example, "there is a low speed vehicle on the right lane, so it's better to follow the front vehicle", or, "The right lane is free, and the front vehicle is about to yield the right of way. I can make a slight maneuver to pass quickly". Especially in populated cities with complex traffic conditions, human drivers are even less likely to deliberately think and come up with specific higher-level behaviors, and strictly control the car, as shown in Fig \ref{DT}. Based on the above analysis, driving tendency is defined as a term used to quantify which type of behavior a driver prefers to perform while driving. Autonomous driving can be seen as a policy search problem for a global optimization problem, so the essence of driving tendency is to quickly reduce the search space.

As shown in Fig. \ref{DTexample_a}, at a junction with a green traffic light, the front car with its left-turn signal on and shows a clear left-turn behavior. The driver's goal is to continue moving forward, so the driver will choose a slight rightward maneuver to bypass and then proceed straight along the current lane. Clearly, any leftward driving tendency other than driving in the current lane would be unreasonable.




Fig. \ref{DT_example_b} provides an additional example. In a two-lane road, the ego car travels in the left lane. As the car in the right lane moves slower, it would be illogical to exhibit a rightward driving tendency. In this situation, the driver would choose to reduce the speed and follow the front car in the current lane.


\textbf{\emph{Remark 1}}: \emph{Driving tendency can be thought of as a smooth transition between an infinite number of successive local optimization objectives in a driving task.}

In summary, this paper applies the concept of driving tendency to the proposed algorithm to improve the learning efficiency. The driving tendency is modeled as a deep neural network,  called the  driving tendency network. This network, together with a receding horizon optimization component, constitutes the policy function approximator and is solved with maximizing cumulative return as the optimization objective.

\subsection{Potential Impact Vehicles Selection}

Traffic dynamics is used to model the interaction between the ego vehicle and traffic vehicles\cite{liu2022interaction} (all vehicles on the road except the ego vehicle).  The traffic scenario is considered to include $n+1$ vehicles, where vehicle $0$ is the ego vehicle and the other $n$ vehicles are traffic vehicles. Define the traffic vehicles as the set  $\mathbb{Q}= \left\{ {V_{tra}^1,V_{tra}^2,V_{tra}^3, \cdots ,V_{tra}^n} \right\} $ In particular, the traffic dynamics can be described by the following discrete-time model:

\begin{equation}
\label{traffic_model}
{\bar s_{t + 1}} = f\left( {{{\bar s}_t},{{\bar u}_t}} \right),\
\end{equation}
where ${\bar s_t} = \left( {s_t^0,s_t^1,s_t^2, \ldots ,s_t^n} \right)\  $denotes the traffic state at time $t$, with $s_t^0\ $denotes the state of ego vehicle. $s_t^k,k \in \left\{ {1, \cdots ,n} \right\}\ $denotes the state of $kth$ traffic vehicle; ${\bar u_t} = \left( {u_t^0,u_t^1,u_t^2, \ldots ,u_t^n} \right)\ $denotes the actions of all $n+1$ vehicles at moment $t$.

As shown in Eq. \ref{traffic_model}, the complexity of the traffic dynamics model increases with the number of vehicles considered. As a result, it is challenging to apply mechanistic models for modeling. However, not all surrounding vehicles within the perceptual range will interact directly with the vehicle in a realistic urban autonomous driving process. Without loss of generality, we define potential impact vehicles as the set:

Easy to know  $\Re  \subseteq \ \mathbb{Q}$

\begin{figure}[!t]
\centering
\includegraphics[width=3.5in]{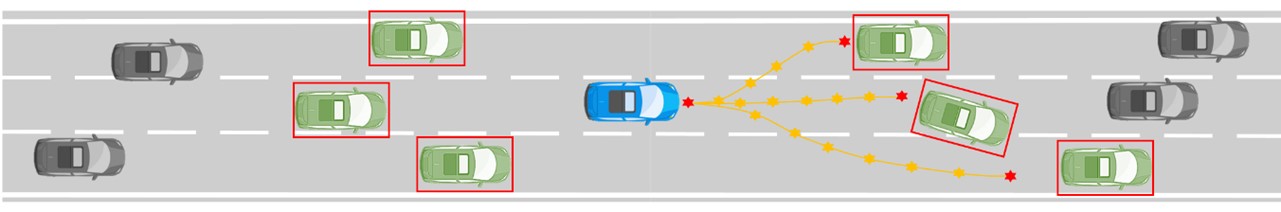}
\caption{Traffic vehicles with potential impact on the ego vehicle.}
\label{strong_interaction}
\end{figure}

\begin{algorithm}[H]
\caption{Traffic interaction model.}\label{alg:alg1}
\begin{algorithmic}
\STATE \textbf{Input:} state input ${s^f}$, HD map info ${H_D}$;
\STATE \textbf{Initialize:} terminal states set $O = \emptyset\ $;
\STATE \textbf{Set parameters:} maximum acceleration $a_{max}$ , acceleration exponent $\delta$, desired speed $v_0$, minimum clearance distance $s_0$ , safe time headway $T$, desired acceleration $b$, prediction horizon $N_p$, time interval $T$;
\STATE \textbf{Filter} potential impact vehicles using Eq. \ref{strong_interaction_selection}: ${\Re ^m} = \left\{ {{V_1},{V_2},...,{V_m}} \right\} = \Theta ({s^f},{H_D})$
\STATE \textbf{Assignment} ${V_i} = \left[ {{x_i},{y_i},{\varphi _i},{v_i}} \right],i = 1,2,...,m$ in ${\Re ^m}$ from $s^f$;
\STATE \hspace{0.5cm}\textbf{for} \emph{i=1, 2, ..., m} \textbf{do}
\STATE \hspace{1.0cm}$D_{x\_1}^i = {x_i} - {x_e}$, $D_{v\_1}^i = {v_i} - {v_e}$, ${v_{e\_1}} = {v_e}$;
\STATE \hspace{1.0cm}\textbf{for} \emph{k=1, 2, ..., $N_p$} \textbf{do}
\STATE \hspace{1.5cm}$a_{IDM}^k = IDM({v_{e\_k}},D_{v\_k}^i,D_{x\_k}^i)$
\STATE \hspace{1.5cm}${v_{e\_k + 1}} = {v_{e\_k}} + a_{IDM}^k \cdot T$
\STATE \hspace{1.5cm}$D_{x\_k + 1}^i = D_{x\_k}^i + {v_i} \cdot T - {v_{e\_k}} \cdot T$
\STATE \hspace{1.5cm}$D_{v\_k + 1}^i = {v_i} - {v_{e\_k + 1}}$
\STATE \hspace{1.0cm}\textbf{end for}
\STATE \hspace{1.0cm}${\tau ^i} = \left[ {D_{x\_{N_p} + 1}^i,{y_i},{\varphi _i}} \right]$
\STATE \hspace{1.0cm}$O \leftarrow O \cup {\tau ^i}$
\STATE \textbf{end for}
\STATE return $O = \left\{ {{\tau ^1},{\tau ^2},...,{\tau ^m}} \right\}$
\end{algorithmic}
\label{alg1}
\end{algorithm}

\begin{equation}
\Re  = \left\{ {V_{tra}^1,V_{tra}^2,V_{tra}^3, \cdots ,V_{tra}^m} \right\}.\
\end{equation}

\textbf{\emph{Remark 2}}: \emph{The set $\complement_\Re\mathbb{Q}$ does not refer to traffic vehicles that have absolutely no interactions with the ego vehicle, but rather to traffic vehicles with weak interactive behaviors. }

In the proposed algorithm, a potential impact vehicles selection policy for straight road scenarios is introduced. As shown in Fig. \ref{strong_interaction}, the blue color represents the ego vehicle, while other colors represent traffic vehicles. In a structured road, the traffic vehicles that have potential impact on the ego car are the $m$ vehicles in the forward and backward lanes that are closest to the projected direction of the ego ($m$ represents the number of lanes), as shown by the green vehicles. As indicated in Remark 2, it is noteworthy that apart from the aforementioned 2$m$ vehicles with significant consideration, there exist additional traffic vehicles that indirectly influence the behavior of the ego vehicle by affecting the aforementioned potential impact vehicles. However, their impact is considered to be relatively negligible. Consequently, even if mechanism models are available for modeling, ensuring the accuracy of such models becomes quite challenging. Therefore, the influence of these additional vehicles has been disregarded.

\subsection{Traffic interaction model}

As described in Remark 2, although some traffic vehicles are close to the ego vehicle, the influence of these vehicles can still be ignored when modelling traffic interactions if the ego vehicle’s algorithm can avoid interactions with them. As shown Fig. \ref{strong_interaction}, taking a straight road as an example, the algorithm proposed is designed with hard constraints to avoid making unreasonable lane changes that would have a direct impact on the oncoming traffic behind. Therefore, the traffic interaction model involved is designed for the $m$ vehicles in front of the ego vehicle.

Define the potential impact vehicles selection model as:

\begin{equation}
\label{strong_interaction_selection}
{\Re ^m} = \Theta ({s^f},{H_D}),\
\end{equation}
where ${H_D}\ $is high definition map information, which contains the road id and the topological connection relationship. The above model filters out the nearest vehicle in all $m$ lanes based on information about the ego vehicle and the surrounding vehicles, as well as information from the high definition map, and stores the vehicle status in the set  ${\Re ^m} = \left\{ {{V_1},{V_2},...,{V_m}} \right\} \subseteq \Re \ $.Where ${V_i} = \left[ {{x_i},{y_i},{\varphi _i}} \right],i = 1,2,...,m\ $.

\begin{table}[!t]
\caption{IDM model parameters\label{tab:IDM model parameterss}}
\centering
\begin{tabular}{cc}
\toprule
Parameters & Symbol \& Value \\
\midrule
Safe time headway & $T = 1$ \\
Maximal acceleration  & ${a_{\max }} = 3$ \\
Acceleration exponent  & $\delta  = 10$ \\
Desired deceleration   & $b = 1.7$ \\
Minimum gap distance & ${S_0} = 20$ \\
\bottomrule
\end{tabular}
\end{table}

The regular Intelligent Driver Model (IDM) model is applied as a traffic interaction model, which has the advantage of a small number of parameters, a clear meaning and a good empirical fit. The mathematical description of the model is \cite{treiber2000congested}:

\begin{equation}
{a_{IDM}} = {a_{\max }}\left[ {1 - {{\left( {\frac{{{v_x}}}{{{v_0}}}} \right)}^\delta } - {{\left( {\frac{{{s^*}({v_x},{D_v})}}{{{D_x}}}} \right)}^2}} \right],\
\end{equation}
where ${a_{IDM}}\ $is desired acceleration, ${a_{\max }}\ $is the maximum acceleration, ${v_x}\ $is the vehicle speed, ${v_0}\ $is the desired speed, $\delta$ is acceleration exponent, ${D_x}$ is the relative distance, ${D_v}$ is relative speed, ${s^*}({v_x},{D_v})$ is the expected following distance.

The expected following distance is expected as:

\begin{equation}
{s^*}({v_x},{D_v}) = {s_0} + \max \left( {0,{v_x}T + \frac{{{v_x}{D_v}}}{{2\sqrt {{a_{\max }}b} }}} \right),\
\end{equation}
where ${s_0}$ is Minimum clearance distance, $T$ is safe time headway, and $b$ is desired deceleration.

The model parameters are listed in Table \ref{tab:IDM model parameterss}. The algorithm pseudo-code of the traffic interaction model is shown in Algorithm 1. The proposed algorithm assumes that all traffic vehicles are travelling at a uniform speed and that the ego vehicle interacts with the traffic vehicles according to the IDM. The inputs to the model are state input ${s^f}$ and high definition map information $H_D$, and the output is the terminal states set $O$. In the second and third lines, $O$ is initialized and the parameters are set. In the fourth line, the potential impact vehicles are filtered through Eq. \ref{strong_interaction_selection} to obtain the set ${\Re ^m}$. Lines 5 to 16 describe the deduction process of the interaction model. The potential impact vehicles in the set ${\Re ^m}$ are extracted and used to deduce future states within the time domain $N_p$, until the terminal state information ${\tau ^i} = \left[ {D_{x\_{N_p} + 1}^i,{y_i},{\varphi _i}} \right]$ are obtained, which are pushed into the terminal state set $O$. The terminal states set $O = \left\{ {{\tau ^1},{\tau ^2},...,{\tau ^m}} \right\}$. $O$ is used to find the appropriate initialization point for the optimization problem. The details of the algorithm are described in Section V.

\section{Actor-critic Method based Safety Evolutionary Algorithm}

The proposed architecture is based on the soft actor citric algorithm \cite{haarnoja1861soft}. In contrast to other RL algorithms, the objective of the soft actor critic algorithm includes not only maximizing the desired cumulative reward, but also maximizing the entropy, as shown in Eq. \ref{max_ent}:

\begin{equation}
\label{max_ent}
{\pi ^ * } = \arg \mathop {\max }\limits_\pi  {_{\left( {{s_t},{a_t}} \right) \sim {\rho _\pi }}}\left[ {\mathop \sum \limits_t r\left( {{s_t},{a_t}} \right) + \alpha {\cal H}\left( {\pi \left( { \cdot \mid {s_t}} \right)} \right)} \right],\,
\end{equation}
where $H$ denotes entropy. By maximizing entropy, the output probabilities of actions can be dispersed as much as possible, enabling the exploration of a wide range of states and avoiding the omission of any useful actions.

Similar to all actor critic algorithms, the proposed architecture consists of two main components: the critical and the policy function approximator. The former aims to evaluate the policy performance based on observations input and the current actions. It achieves this by using a neural network to approximate the value function. The latter combines neural networks with nonlinear optimization methods to generate actions based on the observed information. The objective is to enable the agent to maximize the reward while ensuring safety in the environment interaction.

\subsection{Design of critic function approximator}

In this paper, the MLP network is used as the critic function approximator. Overestimation bias is a key problem in value estimation. This problem may cause the network to overestimate the value of state-action pairs, which may disrupt the learning process and lead to suboptimal policies. The solution involves updating two Q-networks independently, and selecting the minimum output value of the two networks as the final Q-value during each update. This method effectively reduces the bias and variance of the estimates, making the learning of the algorithm more stable.  That is:

\begin{equation}
\begin{split}
J_{Q}(\theta_i ) =& \mathbb{E}_{(s_{t},a_{t},s_{t+1})\sim{D}}[\frac{1}{2}(Q_{\theta_i }(s_{t},a_{t})-(r(s_{t},a_{t})+ \\
                &\gamma V_{\theta_i }(s_{t+1})))^2],i=1,2 ,
\end{split}
\end{equation}
where ${Q_\theta }$ is Q network, $r\left( {{s_t},{a_t}} \right)$ is reward, ${V_\theta }$ is value function and $\gamma$ is discount factor.

\subsection{Design of policy function approximator}

The policy approximator consists of three components, including the traffic interaction model, the driving tendency network, and the optimization problem solver. On the basis of the traffic interaction model established in Section IV, this section combines the advantages of learning algorithm and receding horizon optimization to design the policy function approximator. By constructing a constrained nonlinear optimization problem, the policy learning process can be guaranteed to be collision-free, thus enabling the safe evolution of autonomous driving systems.

\subsubsection{Optimization problem design}

In this paper, the optimization problem is constructed with the idea of model predictive control (MPC). The core of MPC is the receding horizon optimization, whose idea is to repeatedly solve the optimization problem at each sampling moment and output the control action, and timely correct various complex situations in the control process. In this way, despite the limited accuracy of the model in each prediction time domain, the resulting control policy is guaranteed to perform well with the help of feedback. Interestingly, human drivers themselves do not perform accurate modelling and prediction, but rather adopt a mechanism similar to receding horizon optimization, i.e., giving the most sensible action based on the actual state of the moment.

To make the problem tractable, a local solution to the trajectory optimization problem is usually obtained at each step $t$ by estimating the optimal actions at $t+N_p$ on a finite horizon $N_p$ and performing the first action at step $t$. The general form of an optimization problem is defined as follows \cite{gros2019data}:

\begin{equation}
\label{MPC}
\begin{split}
\Pi^{\mathrm{MPC}}\left(s_t^f\right)=& \arg \min _{\mathbf{a}_{t t+N_p}}(V_p(s_{N_p}^f, \theta_\pi)+ \sum_{i=t}^{t+N_p} \gamma_p^i \cdot \\
                                     &\mathcal{L}_p(s_i^f, a_i)),
\end{split}
\end{equation}

\begin{align*}
\hspace{-20em}\text{s.t.} \quad & F(s_i, a_i) = 0 \\
& \Gamma(s_i, a_i) \geq 0,
\end{align*}
where $s_t$ is the state input at time $t$, ${a_{t:t + {N_p}}}$ is the action sequence from step $t$ to step $t+N_p$, ${\gamma _p}$ is the discount factor of MPC, ${V_p}$ is the terminal cost, ${{\cal L}_p}$ is the stage cost, ${\theta _\pi }$ is the optimal parameter, $F( {{s_i},{a_i}})$ and $\Gamma( {{s_i},{a_i}})$ are equation constraints and inequality constraints, respectively. The equation defines a total cost function that includes the generation of a driving tendency network, which considers the overall driving policy and behavior tendencies to simulate the decision-making process of human drivers.

The optimal action sequence can be obtained by solving the above optimization problem with the parameter $\theta$ for the state $s_t$ at step $t$ :

\begin{equation}
{a_t}^*\left( {{s_t},\theta } \right) = \left\{ {{a_t}^*\left( {{s_t},\theta } \right),{a_{t + 1}}^*\left( {{s_t},\theta } \right),...,{a_{t + {N_p}}}^*\left( {{s_t},\theta } \right)} \right\}\
\end{equation}

In this paper, ${\theta _\pi }$ is defined as the parameters of neural network. By updating ${\theta _\pi }$, policy function ${\Pi ^{{\rm{MPC}}}}\left( {{x_t}} \right)$ gradually approximates the optimal policy that maximizes the reward.

\subsubsection{Kinematic Vehicle Model}

In this paper, the kinematic model is used to represent the motion of all vehicles. The kinematic bicycle model is defined by the following set of differential equation:

\begin{equation}
\begin{aligned}
\label{dynamic model}
& x=v \cdot \cos \varphi \\
& y=v \cdot \sin \varphi \\
& \varphi=v \cdot \frac{\tan \left(\delta_f\right)}{L},
\end{aligned}
\end{equation}
where ${\delta _f}$ is the front wheel angle, $L$ is the wheelbase.

The discrete-time \cite{zhou2022interaction} form of model at time step $k$  is:

\begin{equation}
\label{state_transfer}
\zeta_{i+1}=f\left(\zeta_i, a_i\right),
\end{equation}
where ${\zeta _i} = {\left[ {\begin{array}{*{20}{c}}{{x_i}}&{{y_i}}&{{\varphi _i}}\end{array}} \right]^T}$ is the state, and ${a_i} = {\left[ {\begin{array}{*{20}{c}}
v&{{\delta _f}}\end{array}} \right]^T}$ is the input. The Eq. \ref{state_transfer} is calculated with the sampling interval $T$. From Section III, ${\zeta _i}$ can be obtained from the state $s_i^f$, i.e., $\zeta_i \subseteq s_i^f$.

\subsubsection{Cost function design}

Autonomous driving systems are required to be able to reach the destinations safely and comfortably with the highest possible efficiency. The nonlinear programming problem is constructed, and the safety actions can be obtained by solving constrained optimization problems. The optimization problem is specified as:

\begin{equation}
\begin{aligned}
{J_{\pi MPC}} =& \sum\limits_{j = 1}^m {\| {{\zeta _{t + {N_p}|t}} - \zeta _{des,t + {N_p}|t}^j} \|_{{Q_j}}^2}  + \\
               &\sum\limits_{i = t}^{t + {N_p}} {\sum\limits_{p = 1}^{{N_{obj}}} {\| {\frac{1}{{{{( {{x_{e,i|t}} - {x_{p,i|t}}} )}^2} + {{( {{y_{e,i|t}} - {y_{p,i|t}}} )}^2}}}} \|} } _R^2 + \\
               & \sum\limits_{i = t}^{t + {N_c}} {\| {{a_{i|t}}} \|_{{R_u}}^2}  + \sum\limits_{i = t}^{t + {N_c}} {\| {\Delta {a_{i|t}}} \|_{{R_{du}}}^2},
\end{aligned}
\end{equation}
where $N_p$ is control horizon, $N_c$ is prediction horizon, $\left[ {\begin{array}{*{20}{c}}{{x_{p,i|t}}}&{{y_{p,i|t}}}\end{array}} \right]$ are the prediction trajectories of the traffic vehicles in $N_p$ steps, $\left[ {\begin{array}{*{20}{c}}{{x_{e,i|t}}}&{{y_{e,i|t}}}\end{array}} \right]$ is the prediction trajectory of the ego vehicle in $N_p$ steps, and $N_{obj}$ is the number of traffic vehicles.

The terminal states set proposed in Section IV.C is used to build the first term of the cost function, where $\zeta _{des,t + {N_p}|t}^j = {\left[ {\begin{array}{*{20}{c}}
{x_{des,t + {N_p}|t}^j}&{y_{des,t + {N_p}|t}^j}&{\varphi _{des,t + {N_p}|t}^j}
\end{array}} \right]^T} = {\left[ {\begin{array}{*{20}{c}}
{{\kappa ^j}\left[ 0 \right]}&{{\kappa ^j}\left[ 1 \right]}&{{\kappa ^j}\left[ 2 \right]}
\end{array}} \right]^T}$ is the terminal pose after $N_p$ steps, ${\kappa ^j} = O\left[ j \right],j = 1,2,...m$.

$Q_j$, $R$, $R_u$ and $R_{du}$ denote the weight coefficients of each cost function. The driving tendency is used to adjust the weight factor ${Q_j}$, which can be expressed as follows:

\begin{equation}
{Q_j} = \varepsilon \left( {{\theta _\pi }} \right) \cdot {E_3},j = 1,2,...,m,\
\end{equation}
where $\varepsilon$ is defined as driving tendency factor and is the function of the optimal parameter ${\theta _\pi }$; ${E_3}$ is the third order unit matrix.

\begin{figure}[!t]
\centering
\includegraphics[width=3.5in]{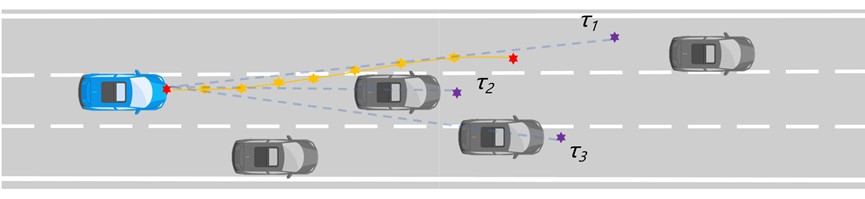}
\caption{Diagram of driving tendencies and optimization trajectory.}
\label{Diagram_of_DT_and_OT}
\end{figure}

The traffic interaction model has been introduced in section III and is used to derive the possible state changes under a specific driving tendency and thus to determine the initial point of the optimization problem solver for the hot start of the optimization algorithm. Let ${\tau _j} = O\left[ j \right],j = 1,2,...,m$, as shown in Fig. \ref{Diagram_of_DT_and_OT}, the purple asterisks are the terminal positions derived from the interaction model, and the yellow asterisks indicate the optimized trajectory under a specific driving tendency factor $\varepsilon$.

\subsubsection{Constraint design}

There are four types of constraints, including vehicle kinematic constraints, control increment constraints, collision constraints, and safety following constraints, defined as follows:

\begin{equation}
\label{dynamic}
\begin{aligned}
{\zeta _{i|t}} = f({\zeta _{i - 1|t}},{a_{i - 1}})\begin{array}{*{20}{c}}
{}&{i = 1,2,...,{N_c} - 1},
\end{array}\
\end{aligned}
\end{equation}

\begin{equation}
\label{action_limit}
\begin{aligned}
\Delta {a_{\min }} < \Delta {a_{i|t}} < \Delta {a_{\max }}\begin{array}{*{20}{c}}
{}&{i = 1,2,...,{N_p}},
\end{array}\
\end{aligned}
\end{equation}

\begin{equation}
\label{collision}
\begin{aligned}
{\left( {{x_{e,i|t}} - {x_{p,i|t}}} \right)^2} +& {\left( {{y_{e,i|t}} - {y_{p,i|t}}} \right)^2} > 4{R^2}\ \\
    &i=1,2,...,N_p,p=1,2,...,N_{obj},
\end{aligned}
\end{equation}

\begin{equation}
\label{safety_distance}
\begin{aligned}
{y_{e,i|t}} < {y_{lead,i|t}} - {d_s}\begin{array}{*{20}{c}}
{}&{i = 1,2,...{N_p}},
\end{array}\
\end{aligned}
\end{equation}
where $\Delta {a_{\min }}$ and $\Delta {a_{\max }}$ are the upper and lower limits of the control increments, ${y_{lead,i|t}}$ are the predicted y-coordinates of the front vehicle of the current lane in the $N_p$ time domain, $d_s$is the safety following distance and $R$ is the safety radius. $R$ can be calculated using the two-circle collision detection model:

\begin{equation}
\label{check_model}
{R^2} = {\left( {\frac{a}{4}} \right)^2} + {\left( {\frac{b}{2}} \right)^2},\
\end{equation}
where $a$ and $b$ are the length and width of the traffic vehicle respectively.

Eq. \ref{dynamic} enforces the functional relationship between the optimization variables satisfies the vehicle kinematics. Eq. \ref{action_limit} enforces the control increments between upper and lower limits to ensure comfort and stability of the ego vehicle. Eq. \ref{collision} and Eq. \ref{safety_distance} enforces that the ego vehicle do not collide with surrounding vehicles to ensure safety.

\subsubsection{Driving Tendency Network Design}

The concept of driving tendency has already been discussed in Section IV. In this paper, a deep neural network is used to fit the optimal driving tendency to output a driving tendency factor $\varepsilon$, and the model input is the environment state information.

In autonomous driving tasks, a reasonable driving preference refers to the safe and efficient expected intentions manifested during the driving process. The actor-critic architecture is a typical RL framework in which the goal of actors is to maximize cumulative return throughout the task. Therefore, under the premise of a well-designed reward function, the optimization goal of the driving tendency network is consistent with the actor function, which is to learn safe and efficient driving behavior by pursuing higher cumulative rewards with the goal of meeting or exceeding human driving performance.

In this paper, the driving tendency factor $\varepsilon$ is defined as a continuous variable and solved by stochastic policy optimization method. In RL community, reparameterization techniques are widely used to improve the stability and efficiency of the learning process \cite{haarnoja1861soft}. $\varepsilon$ can be expressed as:

\begin{equation}
{\varepsilon _t} = {f_{{\theta _\pi }}}\left( {{\tau _t};s_t^\varepsilon } \right) = f_{{\theta _\pi }}^\mu \left( {s_t^\varepsilon } \right) + {\tau _t} \cdot f_{{\theta _\pi }}^\sigma \left( {s_t^\varepsilon } \right)\
\end{equation}
where ${\tau _t}$ is the noise that satisfies the standard normal distribution, $s_t^\varepsilon$ is the state input, $s_t^\varepsilon  \subseteq s_t^f$.
$ {f_{{\theta _\pi }}}\left(  \cdot  \right)$ represents the output of the driving tendency network.

Based on the soft actor critic algorithm, with minimizing KL divergence as the optimization objective, the cost function of the driving tendency network is defined as:

\begin{equation}
\begin{aligned}
\label{plicy_network_loss_function}
J_\varepsilon({\theta _\pi })= &\mathbb{E}_{{s_t^\varepsilon} \sim \mathcal{D}, \tau \sim \mathcal{N}}[\alpha \log \pi_\phi(f_{\theta _\pi }(\tau_t ; {s_t^\varepsilon}) \mid {s_t^\varepsilon})\\
  &-Q_\theta({s_t^\varepsilon}, f_{\theta _\pi }(\tau_t ; {s_t^\varepsilon}))],
\end{aligned}
\end{equation}
where $\alpha$ is temperature hyperparameter, which can be automatically adjusted by formulating a constrained optimization problem. The term $\alpha \log \pi_\phi(f_{\theta _\pi }(\tau_t ; {s_t^\varepsilon}) \mid {s_t^\varepsilon})$ represents the entropy regularization component. This term encourages the policy to explore more diverse actions by maximizing the entropy, thus helping to prevent premature convergence to suboptimal policies.

As described in Section IV, the set ${\Re ^m}$ contains information about the selected $m$ potential impact traffic vehicles. As previously discussed, these $m$ traffic vehicles have a direct influence on the driving behavior of the ego vehicle, so the input state $s_t^\varepsilon$ of the driving tendency network are designed as follows:

\begin{equation}
s_t^\varepsilon  = \left[ {{D_1},{L_1},{D_2},{L_2},...,{D_m},{L_m}} \right],\
\end{equation}
where $D$ and $L$ are the normalized values of the relative longitudinal distance $d$ and the relative lateral distance $l$ between the ego vehicle and the nearest preceding vehicle in the current lane. i.e., $D = \min (\frac{d}{{{D_{\max }}}}1)$ , $L = \frac{{\left( {\min \left( {max\left( {l, - 3} \right),3} \right) + 3} \right)}}{6}$. $D_{max}$ is the maximum longitudinal distance.

In RL, the reward function has a direct impact on the convergence and performance of the algorithm. In order to design a reasonable reward function, a number of important principles need to be followed. Firstly, relying solely on a single positive or negative reward may lead the algorithm into an unreasonable local optimum solution, so both positive and negative rewards should be set and the relationship between reward and punishment should be coordinated. Secondly, the design of the reward function should correspond to the optimization objective of the agent, and there should be appropriate rewards to guide the agent throughout the training process to avoid the problem of reward sparsity.

The reward function required for the driving tendency network training is designed based on the above principles.

Firstly, autonomous vehicles are encouraged to drive within a certain speed range, so speed rewards need to be designed to train the driving tendency network that guides the ego vehicle to pursue as high a speed as possible. The speed reward function is designed based on the principle that the higher the speed of the ego vehicle, the higher the reward received.

\begin{figure}[!t]
\centering
\includegraphics[width=3.0in]{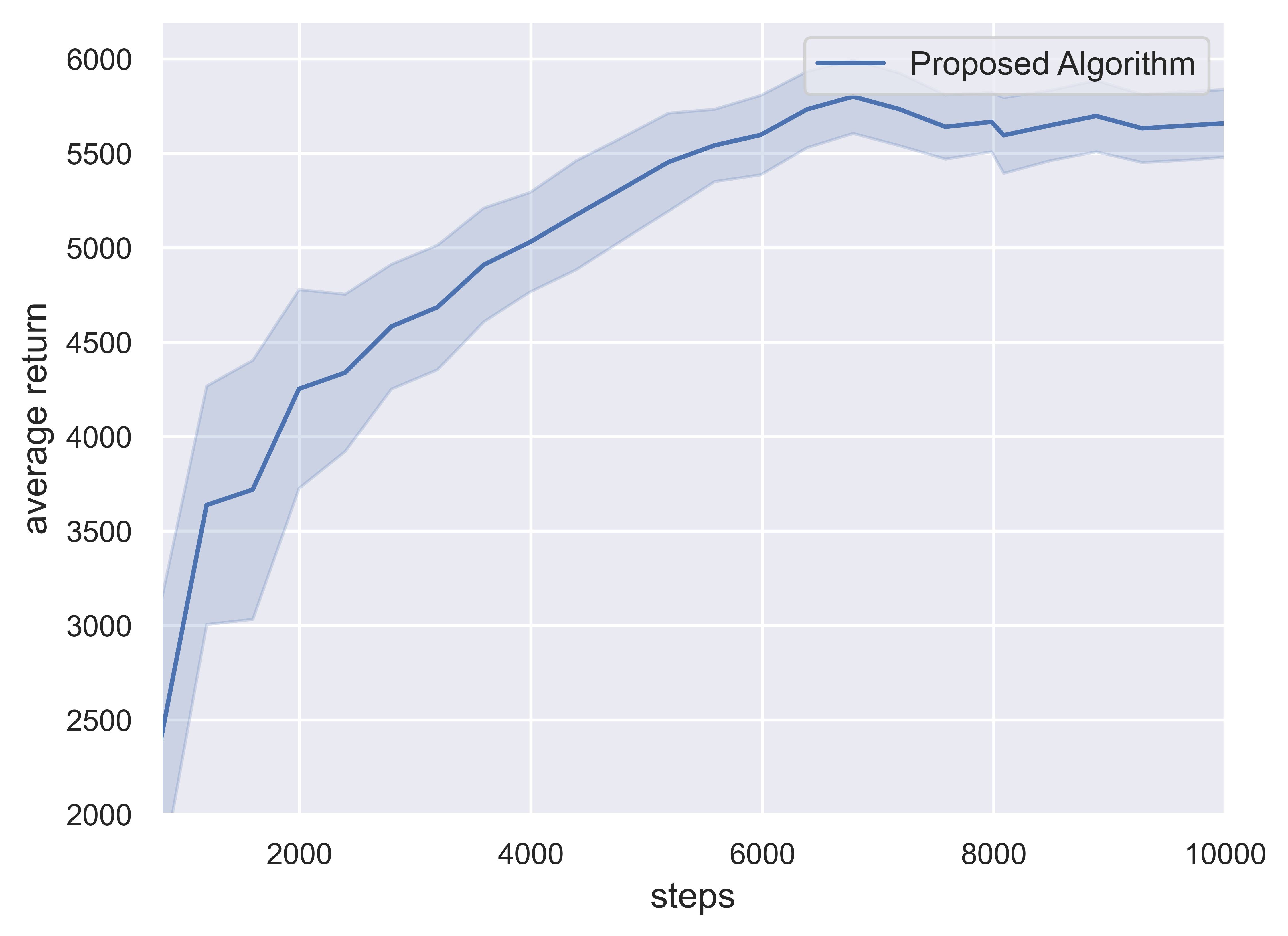}
\caption{ The average return of proposed safe and efficient self-evolving algorithm in 5 runs.}
\label{Learning_curve}
\end{figure}

\begin{figure*}[ht!]
\centering
\begin{tabular}{c}
\includegraphics[width=0.85\textwidth]{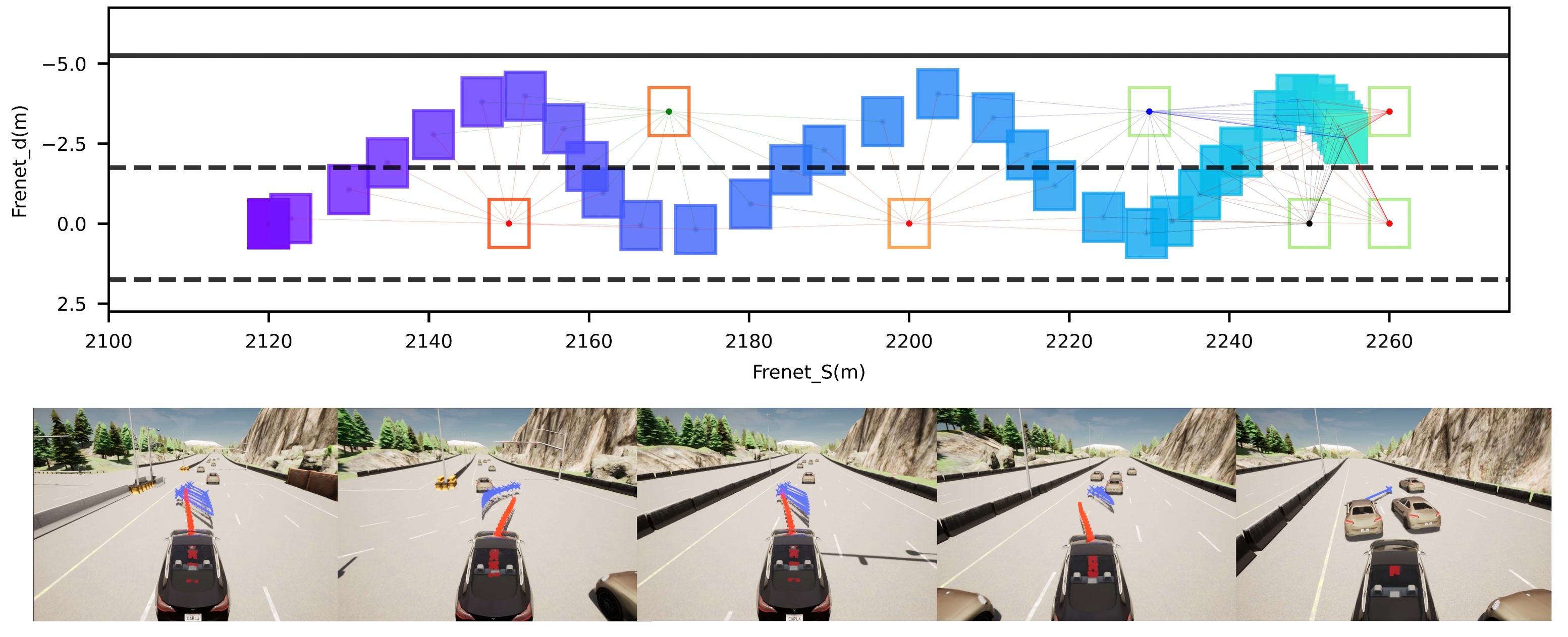}
\end{tabular}
\caption{Visualization of trajectories and driving tendencies of ego vehicles and obstacles in static obstacle scenario.}
\label{traj_plot_twod_rl_mpc_static}
\end{figure*}

\begin{figure*}[!t]
    \centering
    \subfloat[]{\includegraphics[width=0.5\textwidth]{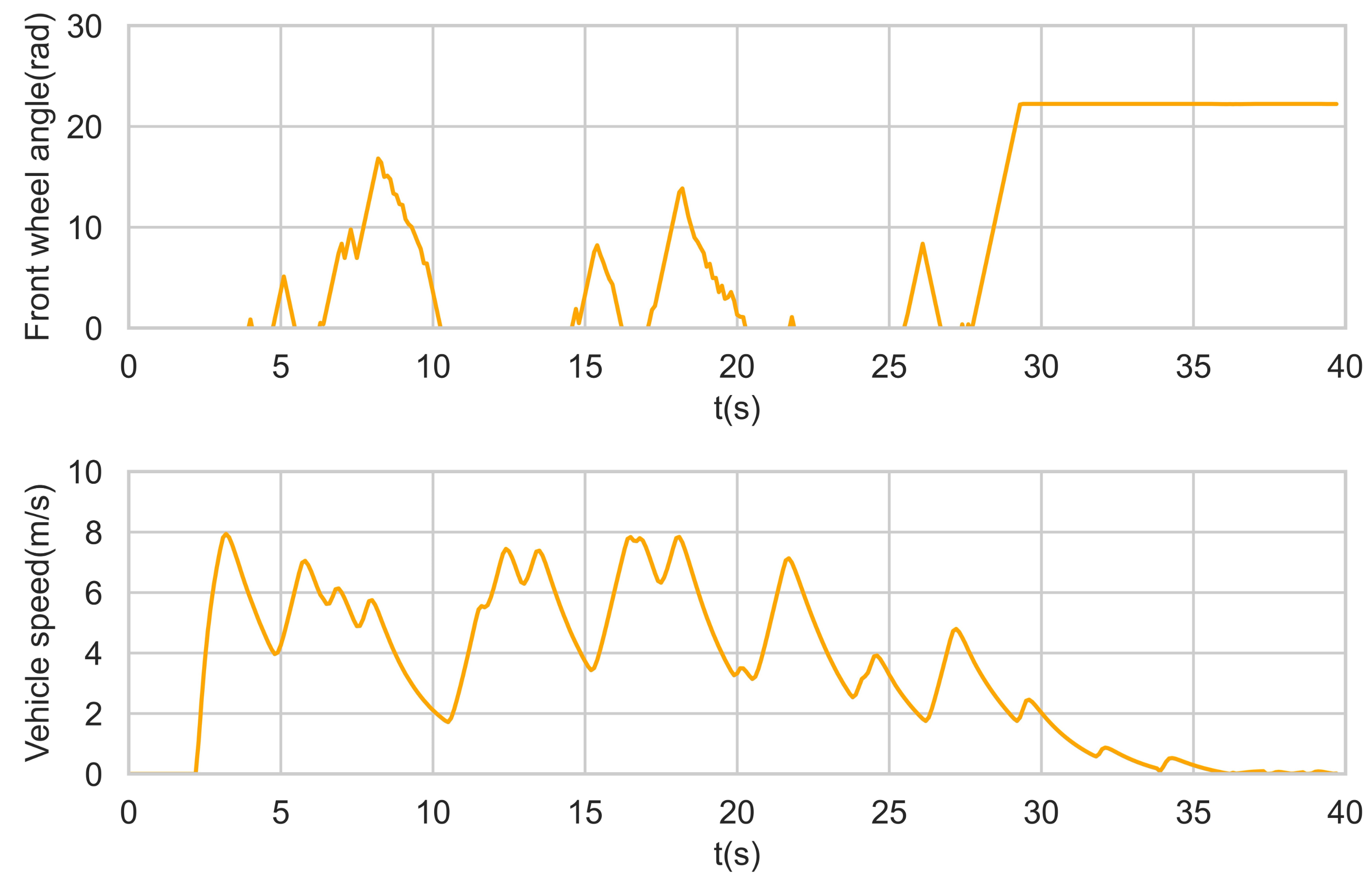}
    \label{vehicle_info_static}}
    \hfil
    \subfloat[]{\includegraphics[width=0.45\textwidth]{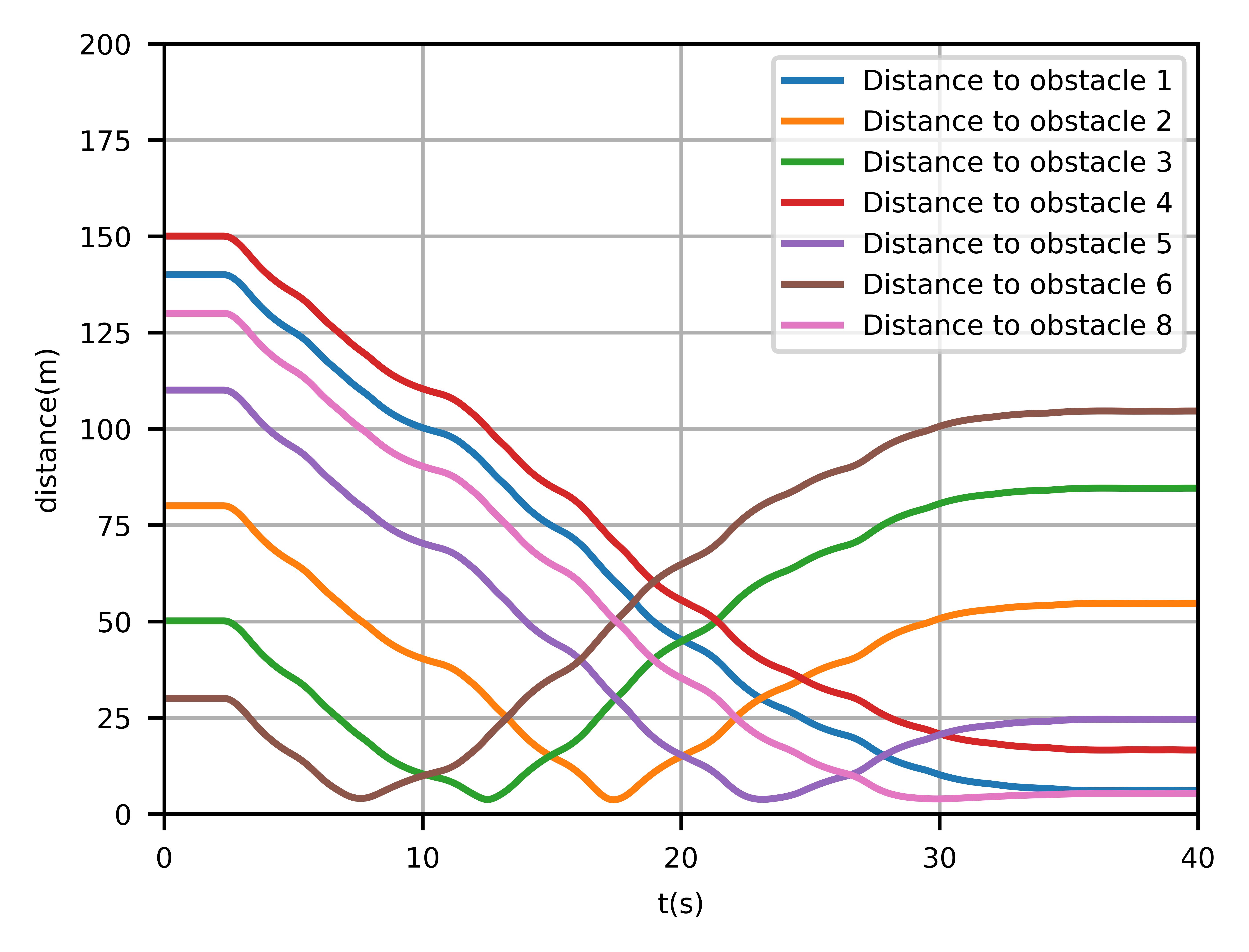}
    \label{Near_dis_static}}
    \hfil
    \caption{Simulation results (a) Steering wheel angle and vehicle speed in the static obstacle scenario (b) Minimum distance between ego vehicle and obstacles in the static scenario.}
\end{figure*}

\begin{figure*}[t!]

\centering
\subfloat[The proposed method]{\includegraphics[width=6.5in]{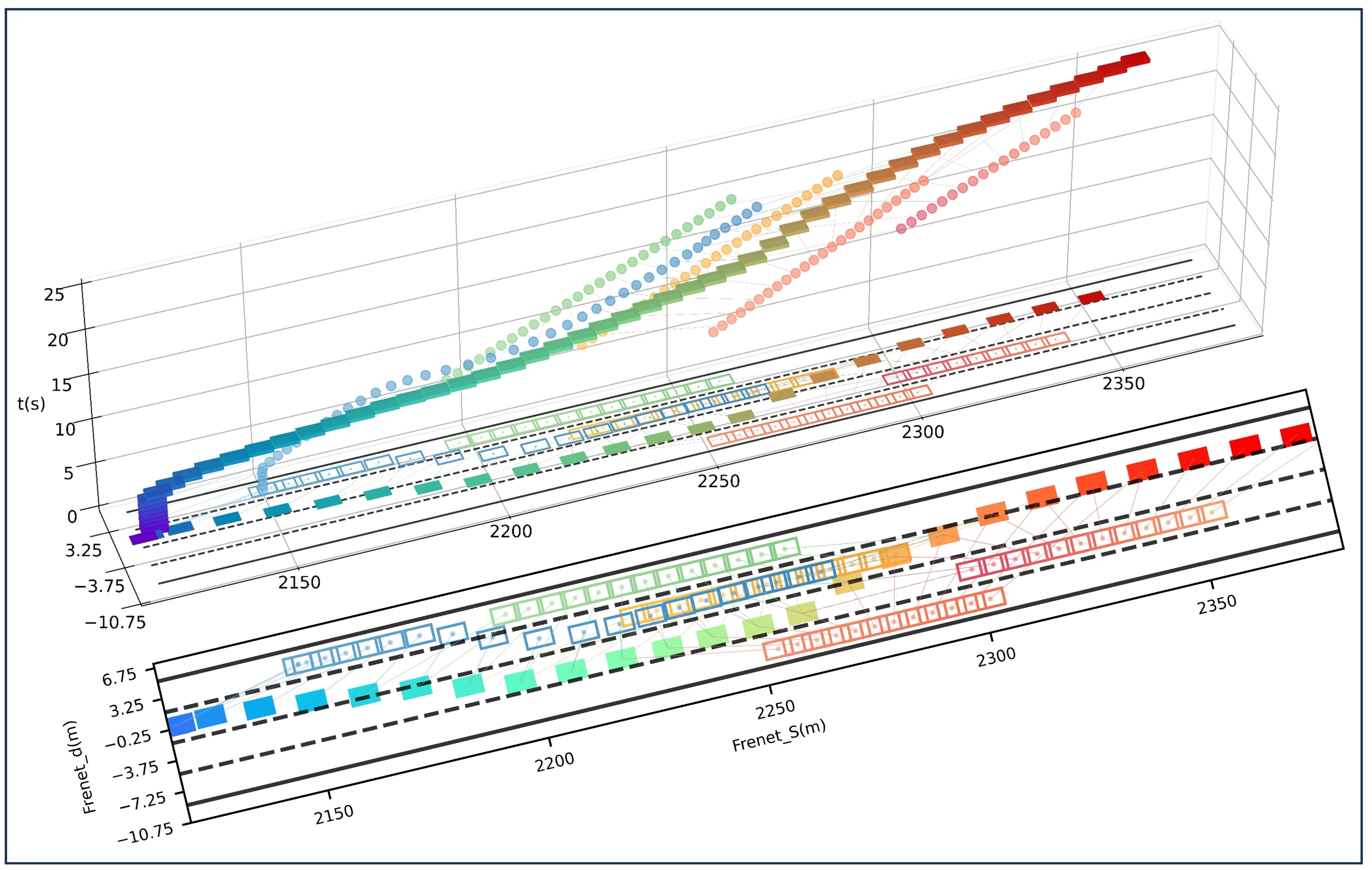}}
\label{The_proposed_method_td}
\hspace{0.1in}
\subfloat[MPC method]{\includegraphics[width=3.3in]{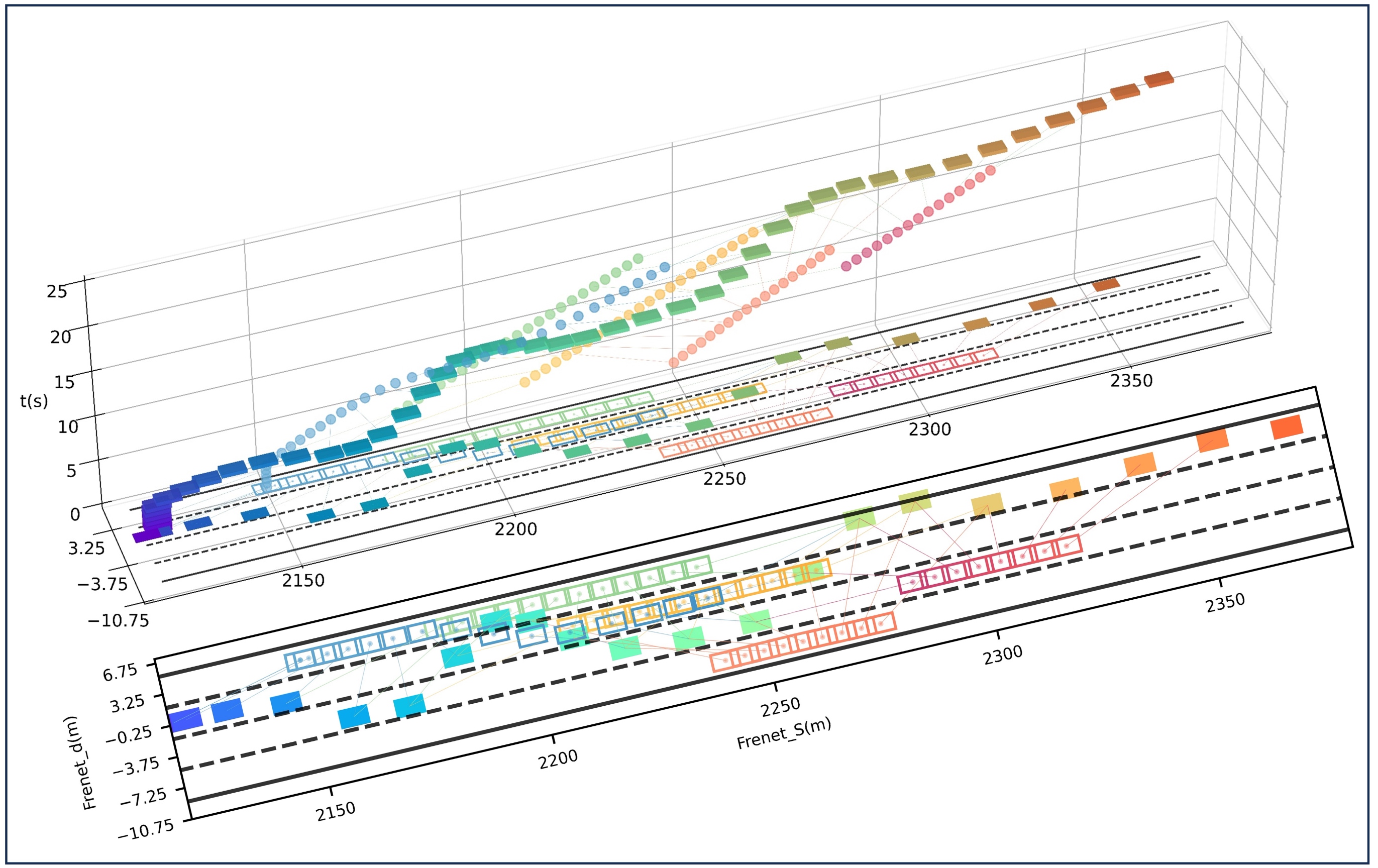}}
\label{MPC_Method_td}
\hspace{0.in}
\subfloat[RL method]{\includegraphics[width=3.3in]{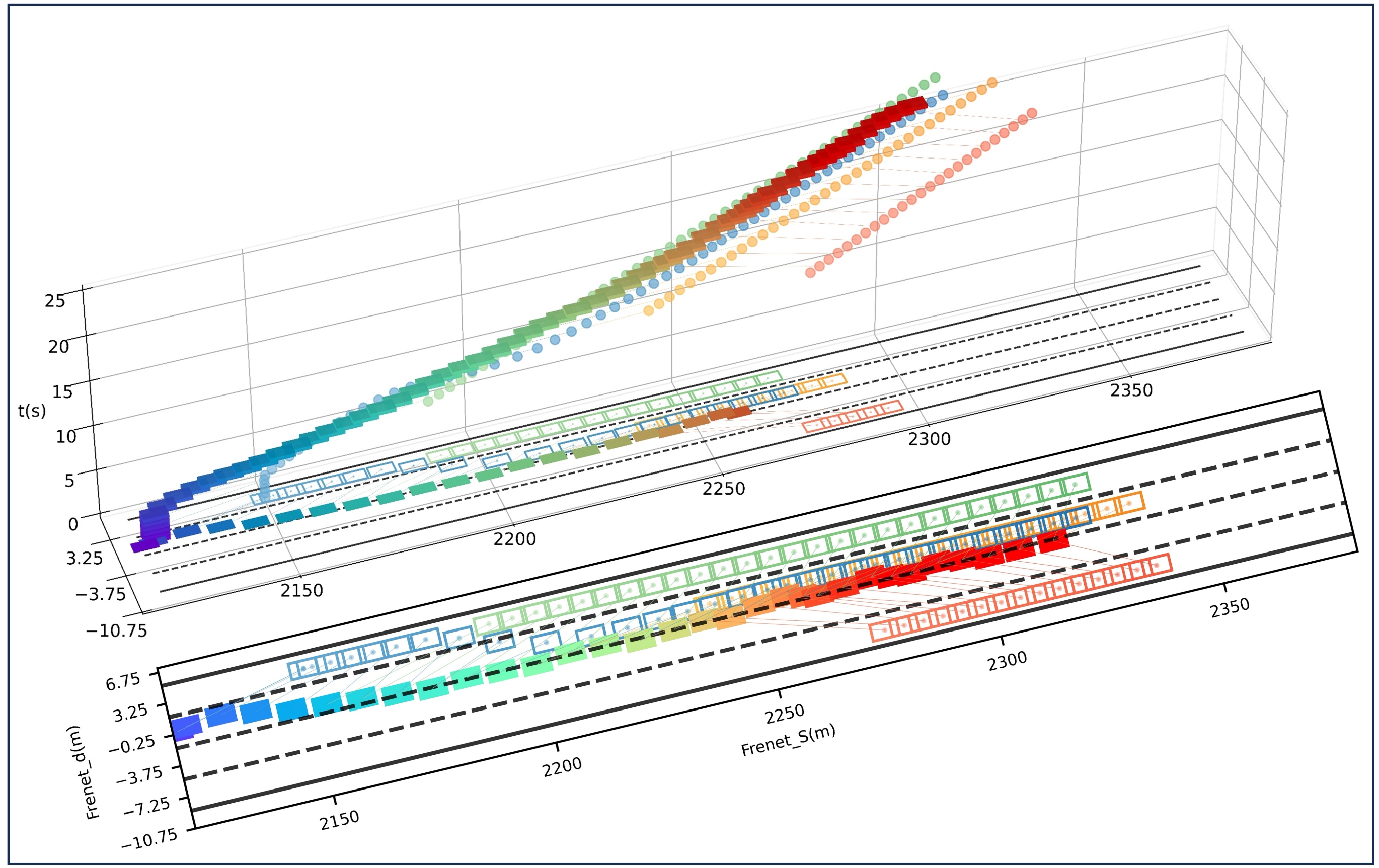}}
\label{RL_Method_td}
\caption{Spatio-temporal diagram of ego vehicle and obstacle trajectories in dynamic traffic flow scenarios. (a) The proposed method. (b) MPC Method. (c) RL method.}
\label{Spatio_temporal}
\end{figure*}

The speed reward is set to:

\begin{equation}
{r_s} = {v_s}\frac{{{\rho _s}}}{{{v_{\max }}}},\
\end{equation}
where ${\rho _s}$ is the speed reward factor, $v_{max}$ is the maximum vehicle speed.

Autonomous vehicles should ensure safety in all scenarios. Therefore, a penalty should be sent when the ego vehicle is too close to the traffic vehicles. Define the safety distance penalty as:

\begin{equation}
{r_n} = \sum\limits_i^m {\frac{{{\rho _n}}}{{\left( {{{\left( {d_s^i} \right)}^2} + {{\left( {d_d^i} \right)}^2}} \right)}}},
\end{equation}
where ${\rho _n}$ is the safety distance penalty factor, $d_s^i$ and $d_d^i$ are the lateral and longitudinal distances between the ego vehicle and the $i_{th}$ traffic vehicles.

To improve training efficiency and prevent reward sparsity problems in training, the driving tendency reward is designed to guide the agent to update gradients in the direction of pursuing more reasonable driving tendency. The driving tendency reward is designed based on the principle that the ego vehicle tends to drive in a more wide-open direction. Using a one-way two-lane road as an example, the driving tendency reward is defined as:

\begin{equation}
{r_{dt}} = \left\{ \begin{array}{l}
 - {\rho _{dt}}{\varepsilon _t} + 10{\rm{  \text{  if  driving tendency  is closed to left}}}\\
{\rho _{dt}}{\varepsilon _t} - 10{\rm{   \text{ if  driving tendency  is closed to right}}},
\end{array} \right.\
\end{equation}
where ${\rho _{dt}}$ is the driving tendency reward factor. The total reward function is:

\begin{equation}
r = {r_s} + {r_n} + {r_{dt}}.\
\end{equation}

The reward function represents the direction in which the agent is encouraged to evolve, with the policy being obtained through the combination of the driving tendency network and the optimization problem solver. All reward coefficients are listed in Table \ref{tab:Reward Coefficients Settings}.

\begin{table}[!t]
\caption{Hyperparameters for the simulation\label{tab:Hyperparameters for the simulation}}
\centering
\begin{tabular}{cc}
\toprule
Parameters & Symbol \& Value \\
\midrule
Simulation step size & $\Delta t = 0.1s$ \\
Discount factor of value network  & $\gamma  = 0.99$ \\
Prediction horizon & $N_p=30$ \\
Control horizon & $N_c=15$ \\
Weight coefficients of cost function U & $R_u = 0.01 $ \\
Weight coefficients of cost function dU & $R_{du} = 0.2 $ \\
Traffic vehicle length & $a=5.0$ \\
Traffic vehicle width & $b=2.0$ \\
Control increment lower limit & $\Delta {v_{\min }} =  - 3m/s,
\Delta {\delta _{\min }} =  - 0.388$ \\
Control increment upper limit & $\Delta {v_{\max }} = 3m/s,\Delta {\delta _{\max }} = 0.388$ \\
Safety following distance & ${d_s} = 5m$ \\
Total training steps & ${T_{total}} = 10000$ \\
Batch size&  ${n_b} = 256$ \\

\bottomrule
\end{tabular}
\end{table}

\begin{table}[!t]
\caption{Reward Coefficients Settings\label{tab:Reward Coefficients Settings}}
\centering
\begin{tabular}{cc}
\toprule
Reward coefficients terms & Symbol \& Value \\
\midrule
High speed  & ${\rho _{{s}}} = 1.3$ \\
Safe relative distance  & ${\rho _{n}} = -0.08$ \\
High value driving tendency  & ${\rho _{dt}} =  20$ \\
\bottomrule
\end{tabular}
\end{table}

\begin{figure*}[t!]
\centering
\subfloat[The proposed method]{\includegraphics[width=6.5in]{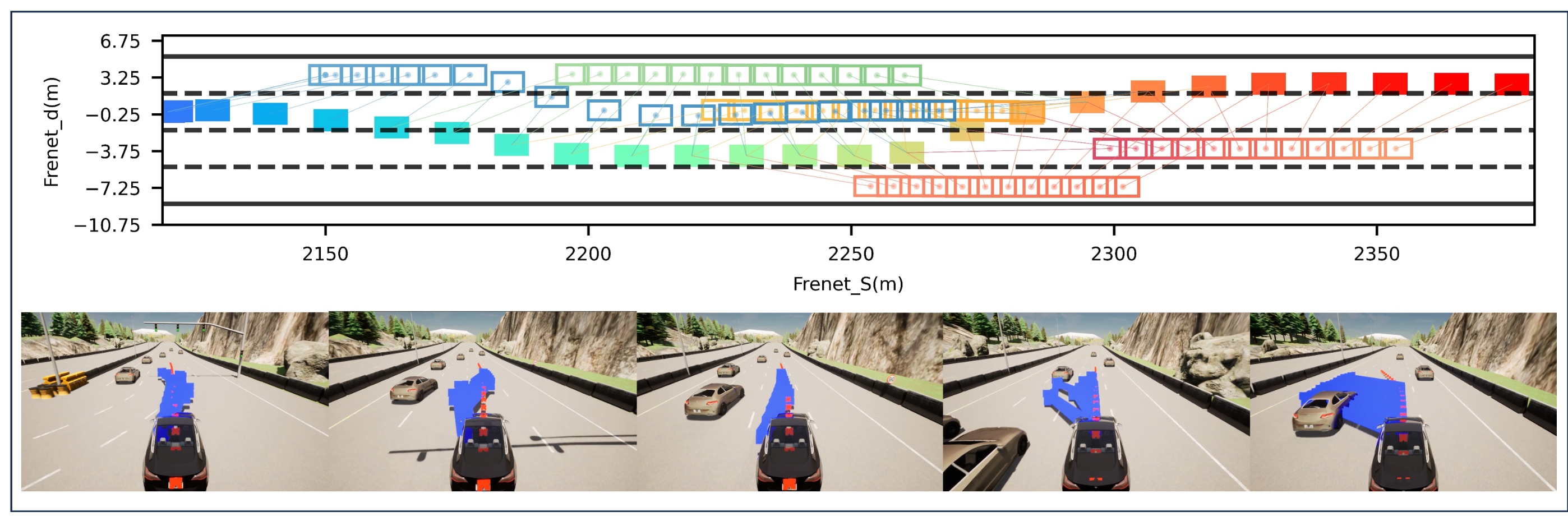}}
\label{The_proposed_method}

\subfloat[MPC method]{\includegraphics[width=3.3in]{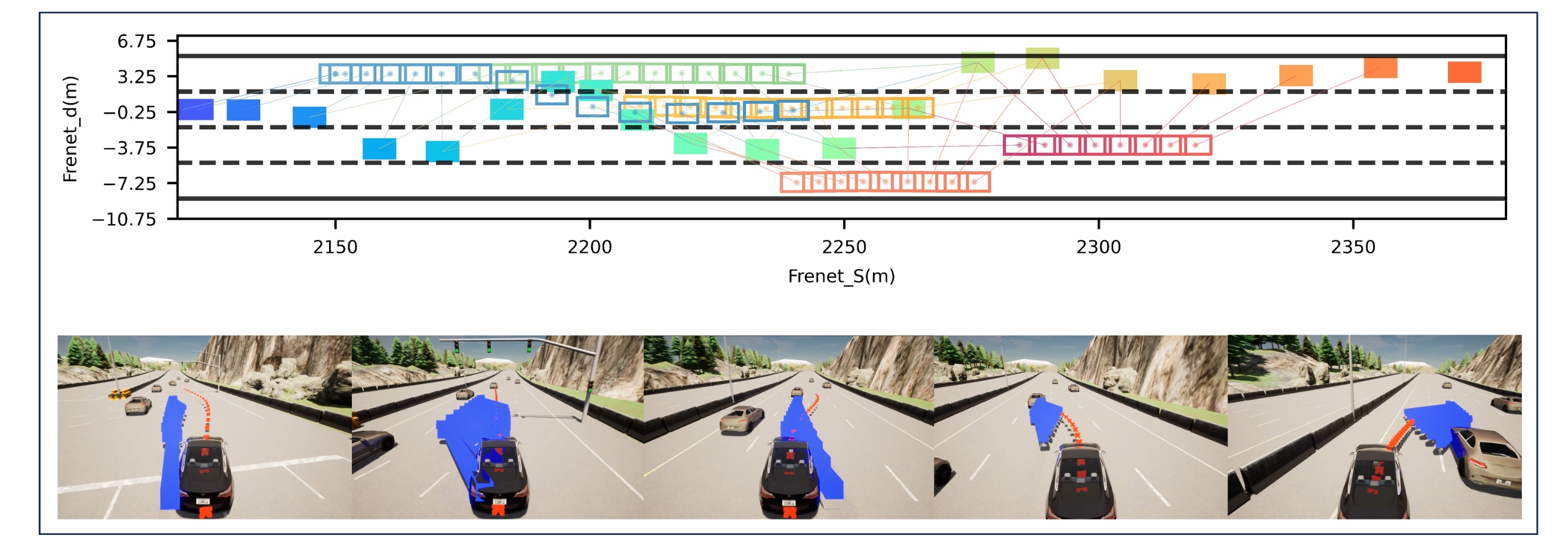}}
\label{MPC_Method}
\vspace{0.in}
\subfloat[RL method]{\includegraphics[width=3.3in]{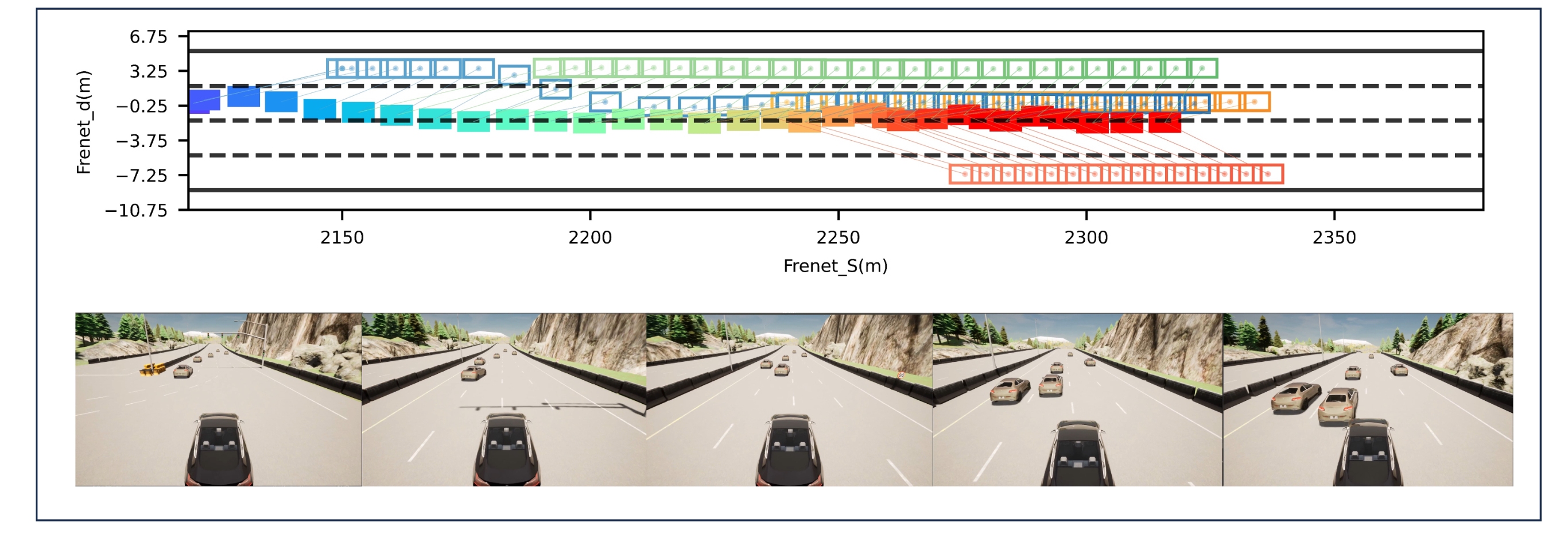}}
\label{RL_Method}
\caption{Visualization of trajectories and driving tendencies of ego vehicles and obstacles in dynamic traffic flow scenarios. (a) The proposed method. (b) MPC Method. (c) RL method.}
\label{traj_twod}
\end{figure*}

\begin{figure*}[!t]
    \centering
    \hfil
    \subfloat[]{\includegraphics[width=0.5\textwidth]{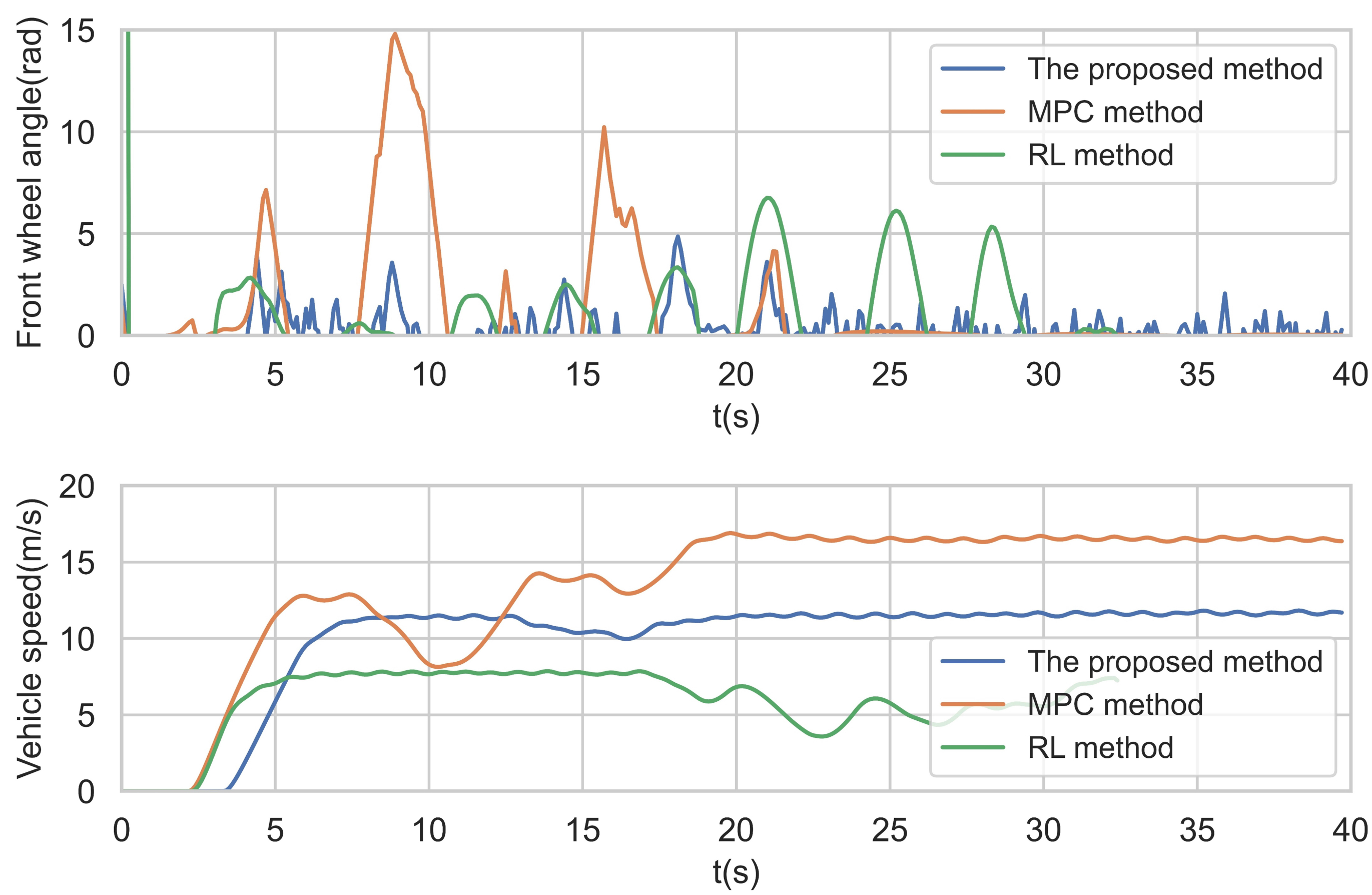}
\label{vehicle_info_mov}}
    \hfil
    \subfloat[]{\includegraphics[width=0.45\textwidth]{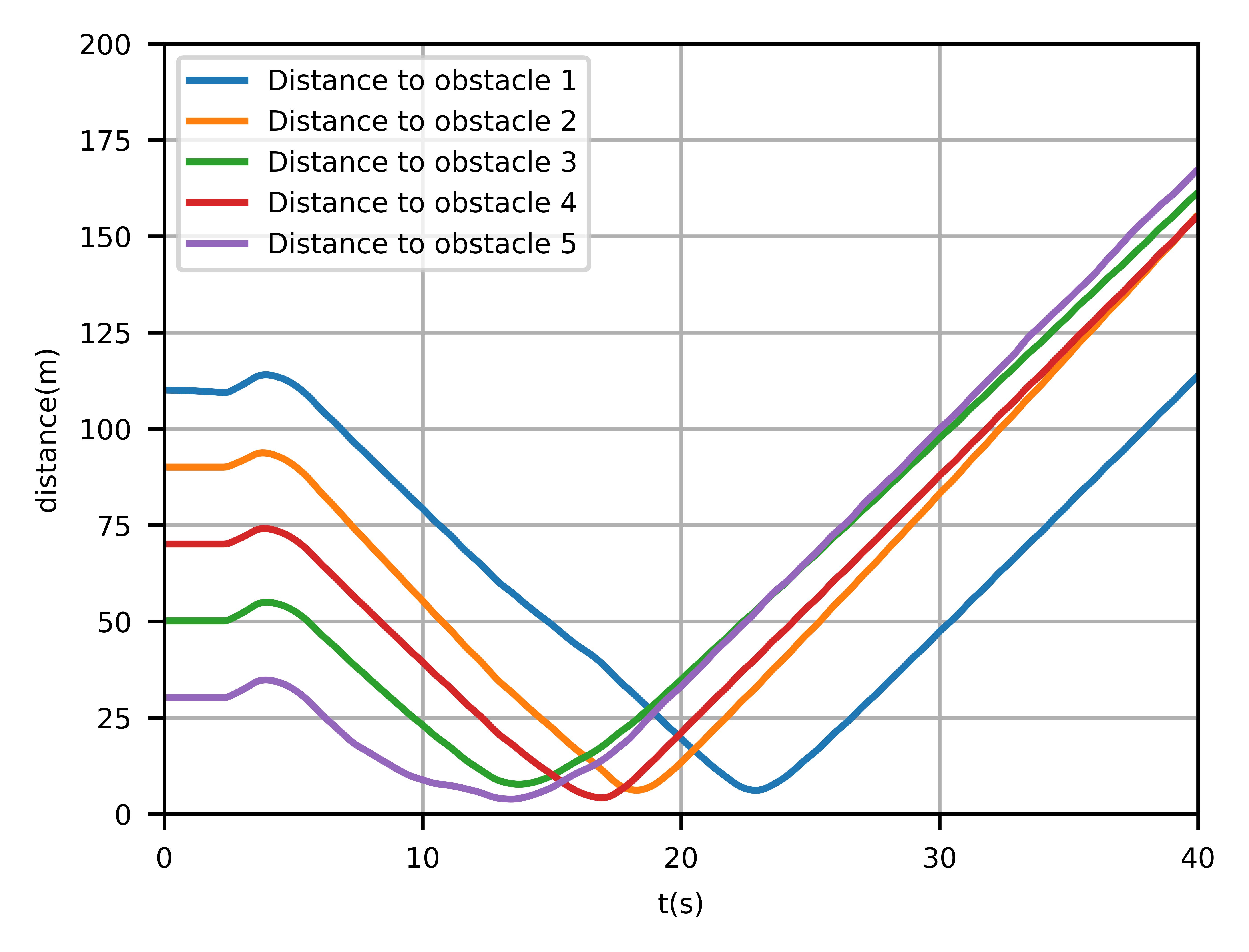}
\label{Near_dis_mov}}
    \caption{Simulation results (a) Steering wheel angle and vehicle speed in the dynamic traffic flow scenarios (b) Minimum distance between ego vehicle and obstacles in the dynamic traffic flow scenarios.}
    \label{info}
\end{figure*}

\begin{table*}[ht]
    \centering
    \begin{tabular}{cccc}
        \toprule
        \textbf{Stats} & \textbf{RL} & \textbf{MPC} & \textbf{Proposed Method} \\
        \midrule
        Collision Rate & 26.7 \% & 3.33 \% & \textbf{0 \%} \\
        Avg. Speed & \(\textbf{11.85} \pm \textbf{0.65}\) $m/s$ & \(8.93 \pm 1.74\) $m/s$ & \(11.20 \pm 0.82\) $m/s$ \\
        Completion Rate & 73.3 \% & 96.67 \% & \textbf{100} \% \\
        \bottomrule
    \end{tabular}
    \caption{The comprehensive performance comparison of RL, MPC and the proposed method after 30 runs.}
    \label{tab:comparison}
\end{table*}

\section{Experiments}

\subsection{Scenario Description}

In this section, the proposed safe and efficient self-evolving algorithm for decision making and control is validated in a dense traffic three-lane lane change scenario. The training environment was built in the simulation software CARLA \cite{moghadam2020end}. The simulation map is chosen as an urban expressway section in Town04, where the lane number is three and the lane width is 3.5$m$. The ego vehicle is chosen as a class B vehicle including a vehicle dynamics model, and the traffic flow model is set to be generated around this vehicle. The maximum time for one round of simulation is 80 $s$. The training scenario is set within a 180$m$ range in front of the ego vehicle, with randomly generated traffic flows in the speed range 6$m/s$ ~12$m/s$.

The parameters of the algorithm are set as shown in Table \ref{tab:Hyperparameters for the simulation}. The value network is defined as a three-layer fully connected neural network with 5 neurons in the input layer, 1 neuron in the output layer and 256 neurons in the hidden layer, and the policy network is defined as a four-layer fully connected neural network with 4 neurons in the input layer, 1 neurons in the output layer and 256 neurons in the hidden layer. In each iteration, ${n_b} = 256$ sets of data is extracted from the replay buffer $D$ for the gradient update of the algorithm.

\subsection{Simulation Results}

The algorithm is deployed in the training environment The computer is equipped with an Intel core i7-10700 CPU, NVIDIA GeForce GTX 1660 SUPER GPU.

As shown in Fig. \ref{Learning_curve}, it shows the average return after 5 runs. It can be seen that the proposed algorithm converges in 4000 to 6000 steps, which indicates that the algorithm has a high learning efficiency. Meanwhile, no collisions occur throughout the training process due to the presence of hard constraints, showing that the proposed framework can achieve safety evolutionary and has the potential for online learning.

In order to fully validate the performance, the experiments are designed to be conducted in static and dynamic scenarios. The algorithm is verified in Frenet coordinate system. This system is widely used in the decision-making and control algorithms due to its ability to correlate actual geometric features and adapt to different road topologies \cite{werling2010optimal}.

\subsubsection{Static Obstacle Scenarios}

The obstacle setup and simulation results are shown in the Fig. \ref{traj_plot_twod_rl_mpc_static}. Frenet\_s and Frenet\_d represent the longitudinal and lateral coordinates in the Frenet coordinate system, respectively. The static obstacles are drawn as rectangles in colors and the ego car is a rectangular box drawn at 0.1s intervals. It presents the ego car achieves obstacle avoidance and reaches the end point. Meanwhile, when the end point is completely blocked, the ego car comes to a reasonable and safe stop. The driving tendency is also shown in the Fig. \ref{traj_plot_twod_rl_mpc_static}, where the blue is the visualized driving tendency output from the network and the orange is the planning trajectory. It shows the proposed algorithm outputs the reasonable driving tendency and the safety planning trajectory.

The Fig. \ref{vehicle_info_static} shows the steering wheel angle and the vehicle speed. It indicates that the method proposed can output smooth and stable control inputs. The minimum distance between the ego vehicle and the obstacles is shown in Fig. \ref{Near_dis_static}. It shows a reasonable and safety distance is maintained between the ego vehicle and the obstacles.

\subsubsection{Dynamic Traffic Flow Scenarios}

Compared to previous scenarios, dynamic scenarios involve complex interactions with surrounding vehicles. We comprehensively conduct comparative experiments to thoroughly validate the effectiveness of the algorithm proposed. MPC algorithm and traditional RL algorithm are chosen for comparison to illustrate the advantages of the proposed algorithm.

The MPC algorithm is constructed using Eq. \ref{dynamic model}-\ref{check_model}, and the optimization objective is defined as maximizing forward speed along the middle lane of while ensuring that the vehicle stays within the road boundaries and avoids collisions. The RL algorithm is chosen as the Soft Actor-Critic algorithm. The reward design includes speed reward, comfort reward, and collision penalty. In order to reduce the search space and improve the algorithm's convergence speed, inspired by paper\cite{moghadam2020end}, the action space $\rm{a}$ of the RL problem is designed as the parameter of the polynomial trajectory.

\begin{equation}
{\rm{a}} = \left[ {{d_s},{a_x}} \right],\
\end{equation}
where $d_s$ is the lateral distance to the endpoint of the local planning trajectory and $a_x$ is the desired longitudinal acceleration. The PID controller is designed to track the desired trajectory, output steering wheel angle and throttle/brake info. In order to facilitate comparison, the parameters of soft-actor-critic algorithm are consistent with the proposed algorithm. The number of training steps is set to 1 million.

The three algorithms are validated in exactly the same dynamic traffic scenarios. The spatio-temporal diagrams of ego vehicle and obstacle trajectories are shown in Fig. \ref{Spatio_temporal}. The traffic vehicles and ego vehicle are represented as rectangular boxes with different colors and are drawn at intervals of 0.1s. As shown in Fig.  \ref{Spatio_temporal}a, it can be seen that the proposed algorithm is able to control the ego vehicle to reasonably accelerate, decelerate, change lanes, and overtake in a traffic environment with consideration of interactions. At t = 5.6$s$, the traffic vehicles in the left lane suddenly choose to change lanes, and the ego vehicle can respond to this situation in time, change lanes to the right safely and smoothly, and overtake all the traffic vehicles in front of it to return to the leftmost lane.

Fig. \ref{Spatio_temporal}b shows the experimental results of applying the MPC algorithm. Under the action of safety constraints, the ego vehicle achieves no collision with the traffic vehicle. However, due to the challenge of constructing a global optimization problem that simultaneously considers both decision-making and control, the optimization algorithm may produce unreasonable continuous lane-changing behavior. From Fig. \ref{Spatio_temporal}a and b, it can be seen that the ablation experiments prove that the driving tendency network proposed in this paper plays a significant role. Fig. \ref{Spatio_temporal}c) shows the experimental results of applying the RL algorithm. In the face of the sudden lane change of the front vehicle, the ego vehicle made a small obstacle avoidance action, but after a period of driving, the ego vehicle still collided with the front vehicle. This shows that the deployed RL algorithm  is hardly able to guarantee absolute safety.

Fig. \ref{traj_twod} provides a more detailed visualization of trajectories and driving tendencies of ego vehicles and obstacles in dynamic traffic flow scenarios. As shown in Fig. \ref{traj_twod}a, the proposed algorithm can still produce reasonable results. Fig. \ref{traj_twod}b and \ref{traj_twod}c depict the actions of the MPC method and the RL method in the same scenario. It can be seen that for the problem of autonomous driving in complex environments, these methods may yield unreasonable actions.

Table. \ref{tab:comparison} shows the comprehensive comparison of RL, MPC and the proposed method after 30 runs. The results show that although the RL method has the highest average speed, it also has a high collision rate. On the contrary, the proposed method achieves 0\% collision rate in complex scenarios including random traffic flow, which means that the safety is guaranteed. At the same time, the average speed has also been significantly improved compared with MPC algorithm. To sum up, compared with the traditional RL algorithm and MPC algorithm, our method not only achieves zero collision in the training process, but also achieves excellent performance in a shorter training time.

Fig. \ref{vehicle_info_mov} shows steering wheel angle and vehicle speed under dynamic obstacle scenario. The front wheel angle curve indicates that the proposed method outputs smaller steering wheel angles compared to the MPC method and RL method, indicating that the proposed method avoids overly aggressive control actions with a smoother speed profile, avoiding abrupt accelerations and decelerations. This demonstrates that the proposed algorithm's performance is well-preserved through the introduction of driving tendencies and the design of control constraints. Fig. \ref{Near_dis_mov} represents the minimum distance between the ego vehicle and dynamic obstacles, showing that the ego car can maintain a reasonable safety distance from other vehicles even in complicated dynamic scenarios.

\section{Conclusion}

This paper proposed a safe and efficient self-evolving algorithm for decision-making and control of autonomous driving systems. The approach treats decision and planning control as a unified optimization problem to be solved, which not only can improve the performance of the evolutionary algorithm quickly and efficiently, but also ensures safety during the learning process. Combining the advantages of rule-based, optimization-based and learning based methods, a hybrid approach based on Mechanical-Experience-Learning is proposed, which can greatly improve the learning efficiency and reduce the training time of the algorithm to ten minutes. The method is validated in a dense traffic three-lane lane change scenario. Experiments show that compared with existing RL methods, the proposed method can ensure safety and collision-free during both training and algorithm deployment processes. Furthermore, by introducing a driving tendency network, it can reduce the search space and achieve more reasonable, safety, and efficient autonomous driving decision and control than MPC method. In future research, we plan to expand the dimension of state space and develop the pedestrian model to improve the ability to deal with complex static scenario and pedestrian interaction. We will also combine domain knowledge and deep learning methods to ensure more explainable effective performance improvements, while enabling safe self-evolution without being too conservative.

\section*{Acknowledgments}
We would like to thank the National Key R\&D Program of China under Grant No2022YFB2502900, National Natural Science Foundation of China (Grant Number: U23B2061), the Fundamental Research Funds for the Central Universities of China, Xiaomi Young Talent Program, and thanks the reviewers for the valuable suggestions.

\bibliographystyle{IEEEtran}
\bibliography{document}

\begin{IEEEbiography}[{\includegraphics[width=1in,height=1.25in,clip,keepaspectratio]{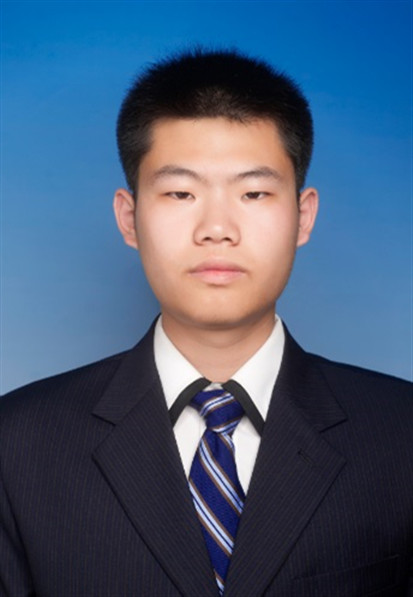}}]{Shuo Yang}
received the B.S. and M.S degree (cum laude) from the College of Automotive Engineering, Jilin University, Changchun, China, in 2017. He is currently pursuing the Ph.D. degree in School of Automotive Studies, Tongji University, Shanghai. His research interests include reinforcement learning,  autonomous vehicle, intelligent transportation system and vehicle dynamics.
\end{IEEEbiography}

\begin{IEEEbiography}[{\includegraphics[width=1in,height=1.25in,clip,keepaspectratio]{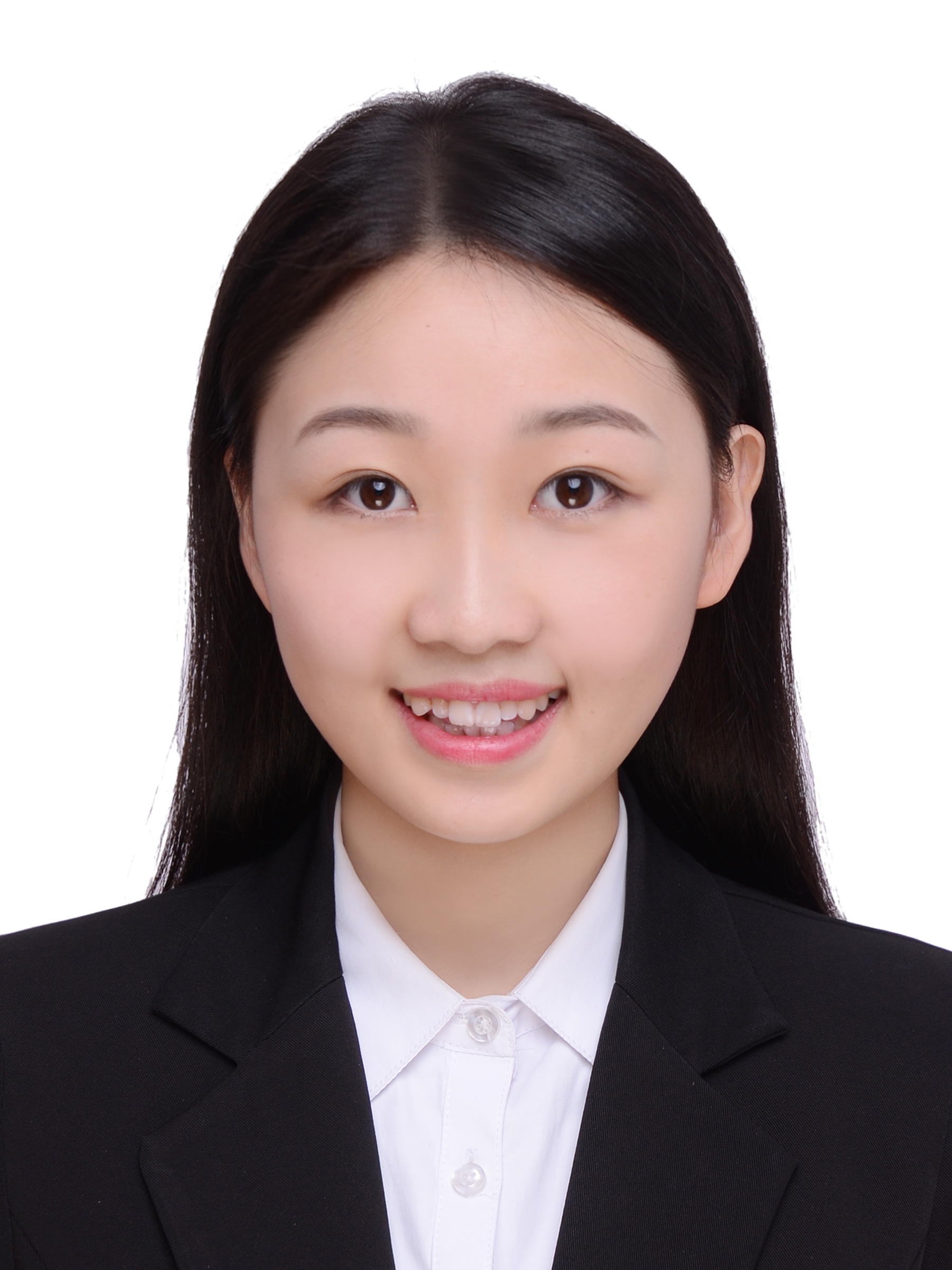}}]{Liwen wang}
received the B.S. degree (cum laude) from the College of Automotive Engineering, Jilin University, Changchun, China, in 2021. She is currently pursuing the M.S. degree in School of Automotive Studies, Tongji University, Shanghai. Her research interests include reinforcement learning, model predictive control and autonomous vehicle.
\end{IEEEbiography}

\begin{IEEEbiography}[{\includegraphics[width=1in,height=1.25in,clip,keepaspectratio]{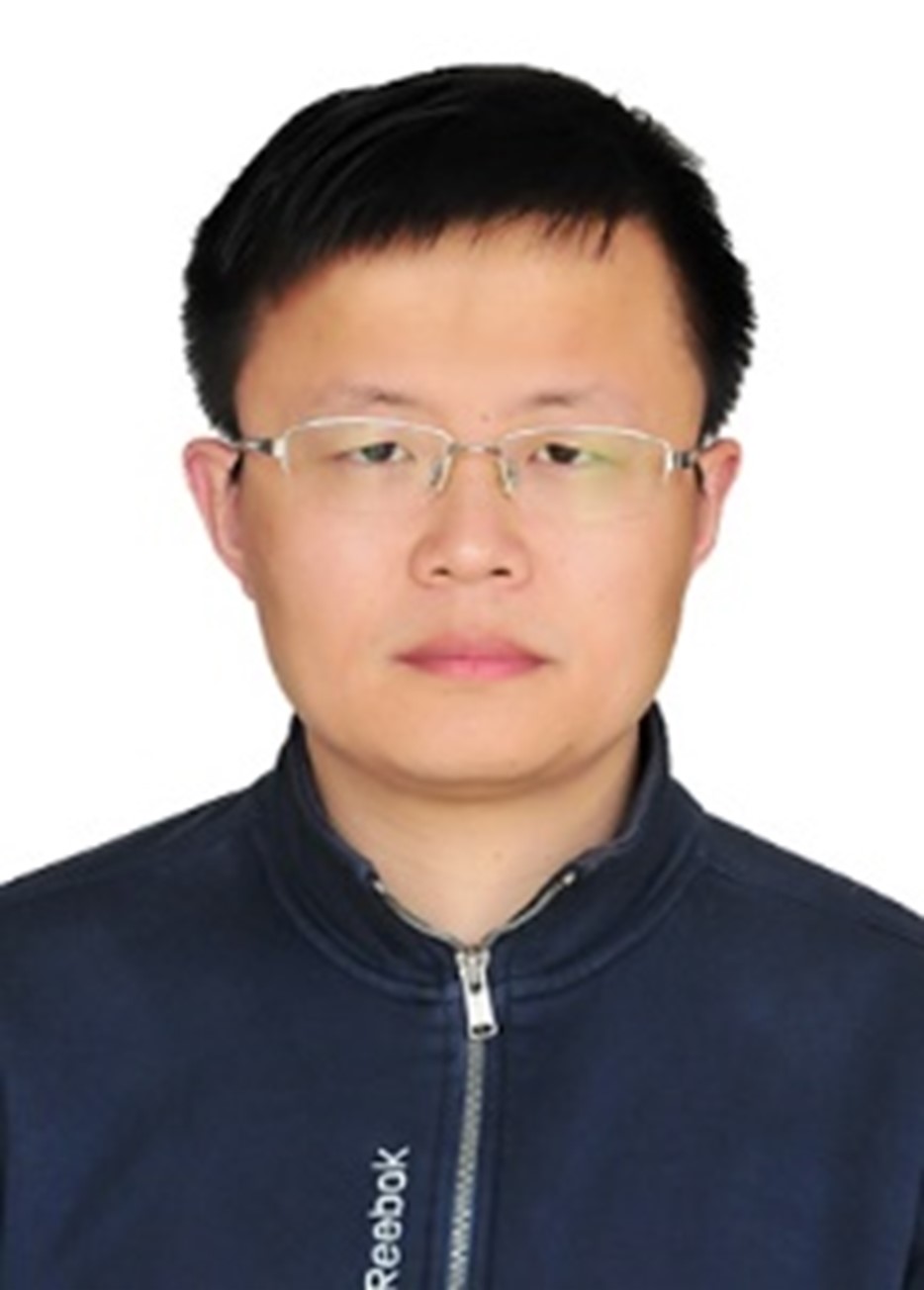}}]{Yanjun Huang}
is a Professor at School of Automotive studies, Tongji University. He received his PhD Degree in 2016 from the Department of Mechanical and Mechatronics Engineering at University of Waterloo. His research interest is mainly on autonomous driving and artificial intelligence in terms of decision-making and planning, motion control, human-machine cooperative driving.

He is the recipient of IEEE VT Society 2019 Best Land Transportation Paper Award. He is serving as AE of IEEE/TITS, P I MECH ENG D-JAUT, etc.
\end{IEEEbiography}

\begin{IEEEbiography}[{\includegraphics[width=1in,height=1.25in,clip,keepaspectratio]{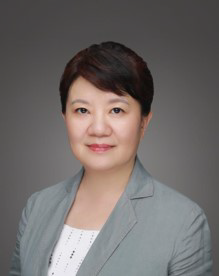}}]{Hong Chen }
(M'02-SM'12-F'22) received the B.S. and M.S. degrees in process control from Zhejiang University, China, in 1983 and 1986, respectively, and the Ph.D. degree in system dynamics and control engineering from the University of Stuttgart, Germany, in 1997. In 1986, she joined Jilin University of Technology, China. From 1993 to 1997, she was a Wissenschaftlicher Mitarbeiter with the Institut fuer Systemdynamik und Regelungstechnik, University of Stuttgart. Since 1999, she has been a professor at Jilin University and hereafter a Tang Aoqing professor. Recently, she joined Tongji University as a distinguished professor. Her current research interests include model predictive control, nonlinear control, artificial intelligence and applications in mechatronic systems e.g. automotive systems.
\end{IEEEbiography}

\end{document}